\documentclass[mnsc,nonblindrev]{informs3}  

\OneAndAHalfSpacedXI

\usepackage[utf8]{inputenc}
\usepackage{url} 
\usepackage{hyperref}
\usepackage{threeparttable}  
\usepackage{multicol}  
\usepackage{multirow} 
\usepackage{graphicx}%
\usepackage{multirow}%
\usepackage{amsmath,amssymb,amsfonts}%
\usepackage{mathrsfs}%
\usepackage[title]{appendix}%
\usepackage{xcolor}%
\usepackage{textcomp}%
\usepackage{manyfoot}%
\usepackage{booktabs}%
\usepackage{algorithm}%
\usepackage{algorithmicx}%
\usepackage{algpseudocode}%
\usepackage{listings}%
\usepackage{fancyhdr}
\usepackage{bm}
\usepackage{tabularx}
\usepackage{booktabs}   
\usepackage{graphicx}   
\usepackage[most]{tcolorbox} 
\usepackage{enumitem}        
\usepackage{lipsum}          

\usepackage{url}

\usepackage{tabularx}
\usepackage{array}
\renewcommand{\arraystretch}{1.2}
\newcolumntype{Y}{>{\raggedright\arraybackslash}X}
 
\usepackage{natbib}
 \bibpunct[, ]{(}{)}{,}{a}{}{,}%
 \def\bibfont{\small}%
 \def\BIBand{and}%

\MANUSCRIPTNO{}

\begin{document}
 
\RUNAUTHOR{Chen, Karaduman, and Kuang}
 
\RUNTITLE{Behavioral Generative Agents for 
Energy Operations}
 
\TITLE{Behavioral Generative Agents for  Energy Operations}
 
\ARTICLEAUTHORS{%
\AUTHOR{Cong Chen,\textsuperscript{a} \"{O}mer Karaduman,\textsuperscript{b} Xu Kuang,\textsuperscript{b}}
\AFF{\textsuperscript{a} Thayer School of Engineering, Dartmouth College, Hanover, NH 03755;\\
\textsuperscript{b} Graduate School of Business, Stanford University, Stanford, CA 94305} 

} 

\ABSTRACT{%
\textbf{{\em Problem definition:}} Accurately modeling consumer behavior in energy operations is challenging due to uncertainty, behavioral heterogeneity, and limited empirical data—particularly in low-frequency, high-impact events. While generative AI trained on large-scale human data offers new opportunities to study decision behavior, its role in operational applications remains unclear. We examine how generative agents can support customer behavior discovery in energy operations, complementing rather than replacing human-based experiments. \textbf{{\em Methodology/results:}} We introduce a novel approach leveraging generative agents—artificial agents powered by large language models—to simulate sequential customer decisions under dynamic electricity prices and outage risks. We find that these agents behave more optimally and rationally in simpler market scenarios, while their performance becomes more variable and suboptimal as task complexity rises. Furthermore, the agents exhibit heterogeneous customer preferences, consistently maintaining distinct, persona-driven reasoning patterns in both operational decisions and textual reasoning. Comparisons with dynamic programming and greedy policy benchmarks show alignment between specific personas and distinct heuristic decision policies. In low-frequency, high-impact events such as blackouts, agents prioritize energy reliability over cost or profit, demonstrating their ability to uncover behavioral patterns beyond the rigidity of traditional mathematical models. \textbf{{\em Managerial Implications:}} Our findings suggest that behavioral generative agents can serve as scalable and flexible tools for studying consumer behavior in energy operations. By enabling controlled experiments across heterogeneous customer types and rare events, these agents can enhance the design of energy management systems and support more informed analysis of energy policies and incentive programs.
}%


\maketitle 

\section{Introduction}\label{sec1}
Energy operations, particularly those targeting demand-side resources such as home batteries, rooftop solar panels, and electric vehicles, are becoming increasingly critical as households adopt new technologies and seek greater efficiency and resilience. Market adoption of energy management systems is projected to surge from \$1.6 billion in 2020 to \$6.5 billion by 2027, driven by smart grid advancements and policy incentives \citep{MarketsandMarkets2021}. These demand-side energy management systems offer significant benefits, including reduced energy costs and improved grid reliability. However, accurately capturing how real customers interact with and manage these technologies remains challenging. Traditional methods such as field experiments can provide valuable insights, but they are often expensive, slow, and difficult to scale. Recent advancements in generative artificial intelligence tools---specifically large language models (LLMs)---offer a promising complementary approach. In this paper, we demonstrate how generative agents (i.e., agents powered by state-of-the-art LLMs) can simulate customer behavior in dynamic energy environments, providing credible and responsive insights without the need for extensive real-world trials.

Our core idea is straightforward: generative agents serve as realistic proxies for human decision-makers in simulated energy scenarios.As emphasized in \cite{park25GenerativeAgents}, the goal of generative agents is not to replace empirical observation but to complement it. These agents facilitate the discovery of customer behavior in energy operations, allowing safe exploration of “what-if” counterfactuals prior to real-world deployment. Because LLMs are trained on vast amounts of human-written text, they naturally capture many aspects of human attitudes, preferences, and reasoning. By carefully prompting these models, we effectively instruct them to role-play diverse customer personas---ranging from a cautious saver to a profit-oriented risk-taker. Embedding these generative agents within simulated energy systems allows us to observe their decision-making processes as they respond naturally to dynamic conditions, such as fluctuating electricity prices and rare but high-impact outages. This approach aligns with the emerging concept of ``homo silicus'' in economic research \citep{FilippasHorton24ECllm}, wherein artificial agents serve as implicit computational representations of human behavior. Unlike traditional energy models that often assume idealized or simplified behavior, generative agents produce decisions that closely mirror actual human reasoning, capturing both heterogeneity and context-dependent preferences.
 
\begin{figure}[htbp]
    \FIGURE
    {\includegraphics[width=0.8\textwidth]{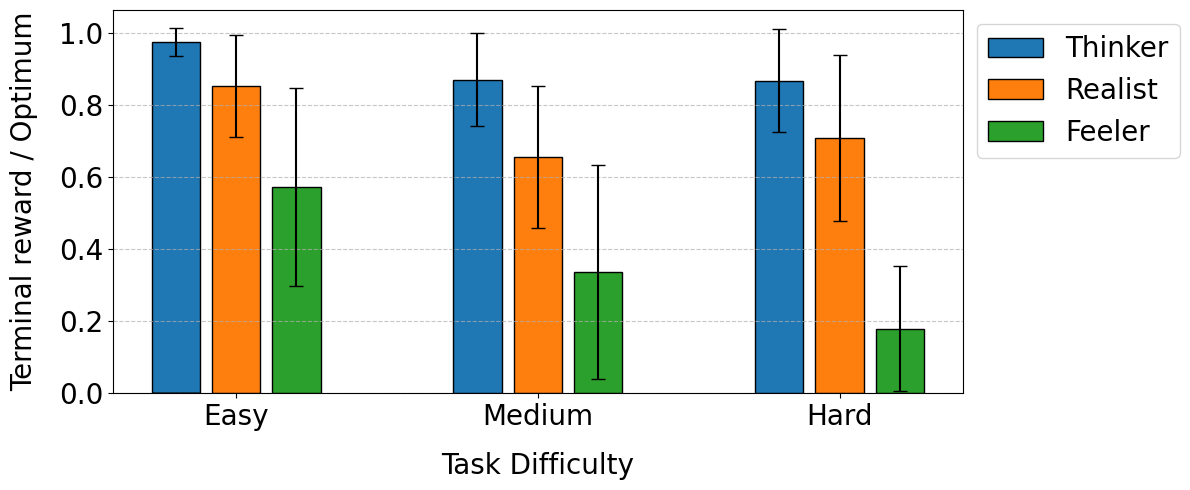}}
    {Terminal rewards of behavioral generative agents. \label{fig:terminal_reward}}
    {This figure compares the average terminal rewards of generative agents with distinct personas—Thinker, Realist, and Feeler—across varying task difficulties (Easy, Medium, Hard). The optimum is given by the Dynamic Programming (DP) policy. Error bars represent standard deviations across simulation runs.}
\end{figure}
Given this setup, an important question arises: how credible and realistic are these generative agents across different electricity market situations? To explore this, we integrate generative agents into a detailed simulation of household energy management, where agents daily decide whether to charge or discharge a home battery based on dynamically changing electricity prices. Figure~\ref{fig:terminal_reward} compares the performance of these generative agents against an optimum benchmark decision policy derived from dynamic programming (DP), across varying task difficulties---Easy, Medium, and Hard. We define task difficulty based on the gap between the DP-optimal reward and the reward achieved by a simple greedy heuristic that charges at low prices and discharges at high prices: scenarios with larger gaps are classified as more difficult. Consistent with realistic human behavior, our agents' decision-making quality varies predictably with task difficulty. In easy price environments, agents, particularly those prompted with a rational \emph{Thinker} persona, closely approximate the DP optimal outcomes, achieving near-optimal performance. However, as tasks become more complex, the agents demonstrate credible human-like limitations: their performance notably declines, and decision variability increases, reflecting increased cognitive demands under uncertainty.

Evaluating agent credibility requires a reference for comparison. Ideally, population-level human behavior provides the ground-truth, but it is difficult and costly to measure, especially for rare, high-impact events such as blackouts. Two common approximations exist: (i) real human data, which captures rich, heterogeneous behavior but is expensive, limited in scale, and often unavailable, and (ii) mathematical benchmarks—such as DP, agent-based models, or game-theoretic approaches—which are interpretable and reproducible but rigid, lacking the ability to reveal unmodeled preferences. In this study, we adopt the latter approach, comparing agent behavior against mathematical benchmarks to evaluate reasoning quality and soundness. By extending the rigidity of math models, generative agents capture diverse behavioral patterns, enabling discovery of plausible decision strategies under conditions such as blackouts. 

Beyond just credibility, generative agents clearly capture diverse customer preferences. Figure~\ref{fig:terminal_reward} also illustrate how prompting agents with distinct personas leads to systematically different outcomes. For example, the cautious \emph{Feeler} persona consistently achieves lower rewards across all difficulty levels due to its preference for intuition-based decisions and maintaining battery reserves. In contrast, the profit-driven \emph{Thinker} persona closely mirrors the optimal DP policy, aggressively pursuing profit-maximizing opportunities. Meanwhile, the \emph{Realist}  persona tends to take a more moderate approach, positioned between these two extremes. These distinct, persona-driven behaviors show that generative agents could represent various consumer segments, making them valuable tools for developing effective energy policies and programs.
 
\begin{figure}[htbp]
    \FIGURE
    {
    \vbox{~~\\
    \makebox[\textwidth][c]{
  \includegraphics[trim={10mm 12mm 0mm 0mm}, clip,width=86mm]{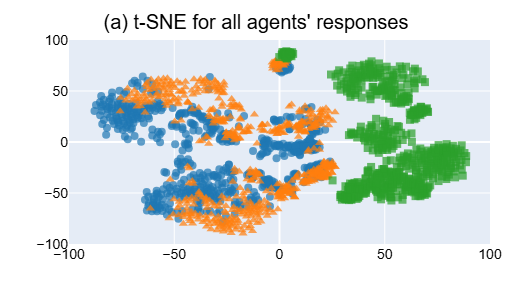}\includegraphics[trim={10mm 12mm 0mm 0mm}, clip, width=86mm]{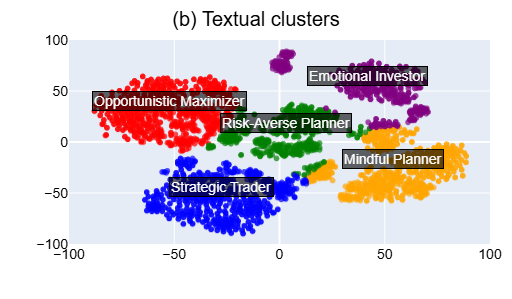}
} 
\makebox[\textwidth][c]{\hspace{18mm}\fcolorbox{gray}{white}{\includegraphics[trim={11mm 7mm 11mm 6.7mm}, clip, width=55mm]{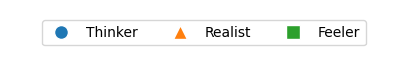}}\hspace{10mm}\fcolorbox{gray}{white}{\includegraphics[trim={100mm 147.7mm 95mm 6.5mm}, clip, width=80mm]{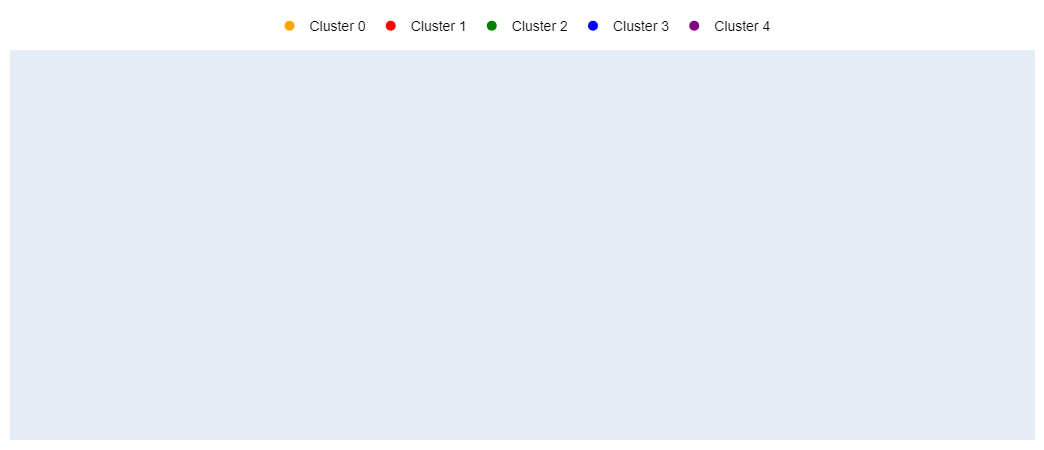}}}}
}
    {t-SNE visualization and preference clusters derived from agents’ textual responses. \label{TR} }
    {(a) t-distributed stochastic neighbor embedding (t-SNE) for all agents' responses. Feeler persona has a clearer response separation from other personas. (b) cluster agents' responses in (a) with distances of textual response embeddings. Preference representations of each cluster are shown in grey shaded labels. }
\end{figure}

Ultimately, the integration of generative agents into energy operations meaningfully bridges the gap between theoretical models and real-world customer behaviors. While traditional energy operation models—such as dynamic programming or heuristic rule-based systems—typically assume simplified, homogeneous customer decisions, generative agents naturally replicate human actions' variability and uncertainty. Furthermore, these agents provide transparency into their decision-making through explicit memory-based reasoning, articulating how they act and the underlying motivations driving those actions. As illustrated in Figure~\ref{TR}, agents’ decision rationales form distinct clusters corresponding to persona-specific preferences, ranging from an “Opportunistic Maximizer,” who emphasizes immediate profitability, to a “Risk-Averse Planner,” who prioritizes security and resilience. In contrast, preferences of the \emph{Feeler} persona center on intuitions, including the “Emotional Investor,” guided by gut feeling, and the “Mindful Planner,” who considers future possibilities with deliberate reserve. These insights into customer decision logic can improve how energy programs and policies are designed and targeted.

To further highlight the practical relevance of generative agents, we examine their responses to rare but critical situations, such as power blackouts. Such scenarios are increasingly significant given recent rises in extreme weather events and grid instability \citep{Sonnen25,Lastbuld23}. As detailed in Section~\ref{sec:results}, generative agents adjust their battery operations quickly and effectively during blackouts, clearly prioritizing household energy security. Their dynamic and clear reasoning provides valuable insights into how real customers might respond under similar uncertain conditions. Overall, our approach demonstrates how generative agents can help anticipate realistic customer behaviors, offering new possibilities for enhancing energy management strategies and policy-making decisions.

\section{Related work}

Growing interest in AI agents for studying human decision-making spans behavioral science \citep{davis24MSOMMLBehavior}, operations management \citep{ chen25MSOMChatGPTBias}, and learning-by-doing experiments \citep{bolton08learning, Liang24NIPS}. A special recent sub-stream,  behavioral generative agents, particularly those employing LLMs, have attracted attention across diverse domains,  including education \citep{bastani24harm}, AI-driven pricing \citep{Cohen25AIpricing}, digital environments \citep{vezhnevets23GABM},  emergent behavior analysis \citep{park23generative}, and biased decision-making \citep{chen25MSOMChatGPTBias}. Our study contributes to this growing literature by examining the use of behavioral generative agents in dynamic energy management simulations. In this section, we review related work in two streams: (i) LLM-based agents as homo silicus for modeling human decision-making, and (ii) approaches to modeling human behavior and preferences in stochastic energy operations.

\subsection{LLMs as Homo Silicus in Decision-Making} 
Recent studies suggest that  LLMs can function as implicit computational representations of human decision-making, a concept termed “Homo Silicus." This approach allows behavioral experiments in silico, facilitating hypothesis generation before real-world testing. \citet{FilippasHorton24ECllm} argue that analogous to economists’ theoretical homo economicus, LLMs can be endowed with information and preferences, enabling simulation and observation of decision-making patterns. Initial experiments using GPT-based agents successfully replicated classic human behavioral biases, such as status quo bias and ultimatum game responses, rather than purely rational outcomes.

Generative agents, integrating memory and planning modules within LLM architectures, further enhance the capability of agents in simulating complex, temporally extended human behaviors
\citep{park23generative}. Unlike traditional rule-based agents or optimization-driven models, generative agent relies on natural language prompts and agent memory to produce contextually appropriate behaviors \citep{vezhnevets23GABM}. For instance, Google DeepMind’s Concordia framework demonstrates that agents can generate actions based on language prompts without explicit utility maximization, yielding more human-like behaviors, though not necessarily optimal. Researchers have also shown that, by conditioning LLM outputs on the features of specific subpopulations, agents can effectively reflect those groups' beliefs and provide predictions with fidelity \citep{hewitt24socialsciencellm, shanahan23NatureRoleplaLLM, kosinski24evaluating, LudwigSendhilAshesh25NBEReconLLM}.

However, significant limitations remain evident in current research about the credibility and robustness of generative behavioral agents in practical experiments. \citet{polachek24homoNot} note that LLMs reliably replicate human behavior only under specific conditions, with deviations becoming prominent in complex environments. Comparative analyses further suggest that, although LLM agents approximate optimal solutions through in-context learning, they display higher variability and lower efficiency than specialized algorithms \citep{vezhnevets23GABM}. \citet{ross24MITllm} and \citet{harding24AIcannot} emphasize that current LLM behaviors neither fully align with human nor purely rational economic benchmarks, indicating that LLM agents are not yet suitable as universal substitutes for human participants.

Overall, it remains unclear when and why agents can generate credible, diverse, and robust behaviors in practical experiments. A key barrier is the lack of reliable benchmarks. Validation can rely on either real human data or mathematical benchmarks, each with strengths and limitations. Real human data provide the most direct measure of behavioral realism, capturing the richness and heterogeneity of actual decision-making \citep{park25GenerativeAgents}. However, collecting such data is often costly, limited in scale, and may be sparse for rare, high-impact events. Mathematical benchmarks, such as agent-based modeling, customer utility optimization, game-theoretic solutions, or heuristic policies, offer precise and reproducible references for rational or near-optimal behavior \citep{Liang24NIPS}, enabling controlled evaluation of agent performance, but they may fail to capture unmodeled, nuanced human decision patterns.

Our research addresses this gap by replicating dynamic and strategic human decision-making in a proof-of-concept energy management experiment. We compare agent decisions with classical operations research approaches and heuristic policies to showcase agents’ capacity to emulate heterogeneous personas believably. We further employ quantitative and qualitative analyses to interpret the reasoning behind agent behavior, providing actionable insights and establishing a foundation for more realistic, data-intensive studies in the future. 
 
\subsection{Modeling Human Behavior in dynamic energy operations} 

Modern energy systems face significant uncertainties from climate change, renewable energy variability, dynamic electricity pricing, unforeseen outage disruptions, and customer behaviors. There's plenty of research about energy operations under uncertainty with methods like dynamic programming \citep{JiangPowell15Storage, Wu23MSOMEnergyStorage, Guajardo24RentToOwn}, chance constraints \citep{ GengXie19AC, YangWu21TPS}, robust optimization \citep{LorcaSunLitvinovZheng16OR_RO, LauingerKuhn24MSOMv2g, yang25MSOMRobustMaintenance} and game-theoretic approaches \citep{Cohen16Green}. Some researchers achieve efficient computation methods in real-time large-scale energy operations with stochasticity from renewable and electricity prices \citep{JiangPowell15Storage, YangWu21TPS}. Nonetheless, dynamic energy operations remain particularly challenging when incorporating uncertain and heterogeneous consumer responses. 

Traditional approaches to modeling consumer decision-making in energy operations typically rely on either direct human-subject experiments \citep{abrahamse05reviewhouseholdEnergy, Harding17TimeVarying, brandon19NaturalExperimentElectricityMarket, Wu22MSOMev} or computational simulations based on idealized rational actor assumptions. Computational models often optimize consumer utility or cost functions \citep{agrawal22MS-drr, abate24ORSDynamicPricing, yao08ORgame, gzuguang24MSOM} or demand curve \citep{kok18MSprice, Cohen16Green}, subject to budget, comfort, and operational constraints. Because accurately modeling customer preferences and learning utility functions can be challenging, another line of computational work employs heuristic rules, such as shifting flexible electricity demand, to guide energy management decisions \citep{chen24CDCdr, mehrabi24DR}. Despite these efforts, real consumer behavior often deviates from rational or rule-based assumptions.

Empirical studies highlight substantial deviations from purely rational behavior, revealing bounded rationality, cognitive biases, and diverse consumer preferences \citep{gillan17HassDP, sloot23HassBehavioralDR}. For instance, consumers frequently demonstrate inertia or prioritize convenience over economic benefits, resulting in inconsistent responses to dynamic pricing and demand response programs \citep{sloot23HassBehavioralDR}. These findings underscore the gap between traditional analytical or heuristic models and actual human decision-making, motivating the search for methods that better capture behavioral realism.

Energy operations face a fundamental trade-off between realism—accurately representing complex human decision processes—and tractability, ensuring models remain solvable and effectively calibrated. Addressing this balance aligns with the broader trend in management science toward behavioral realism, which focuses on how human behavior influences operational outcomes \citep{donohue20MSOMbehavioral}. Our study tackles this trade-off by incorporating generative behavioral agents into dynamic energy simulations. These agents produce sequential decisions and textual responses that reveal implicit, diverse, persona-driven preferences often overlooked by conventional analytical models.  Specifically, we evaluate agents’ behavior change across both routine and emergent operational conditions, which are usually challenging for traditional management science due to limited historical data. By doing so, we can assess whether behavioral generative agents provide credible, diverse, and interpretable simulations of human behavior and support the design of energy management strategies and energy policies.  
\section{Method}\label{sec2}
We propose a behavioral generative agent framework designed specifically for dynamic customer decision-making in energy operations. While most traditional energy operation studies focus on static or one-off experiments, our framework emphasizes continuous, sequential consumer interactions. Each generative agent incorporates structured prompting, reflection, and memory modules to maintain decision consistency and realism over time, minimizing common issues such as inconsistent or unrealistic behavior (often called “hallucinations”). Our agents are embedded within a dynamic battery management simulation, detailed in Section~\ref{Method:TARJ}. A notable benefit of this design is the availability of dynamic programming (DP) as an optimal benchmark, representing an ideal rational decision-maker. We describe this DP benchmark and a simpler, intuitive greedy heuristic policy in Section~\ref{Method:DP}. After generating comprehensive agent data through Monte Carlo simulations, we use quantitative and qualitative analyses (Section~\ref{Method:Quant}) to evaluate agent performance and facilitate behavioral discovery.

\subsection{Behavioral generative agent and the experiment design} \label{Method:TARJ}

\begin{figure}[htbp]
    \FIGURE
    {\includegraphics[trim={0mm 0mm 0mm -3mm}, clip,width=160mm]{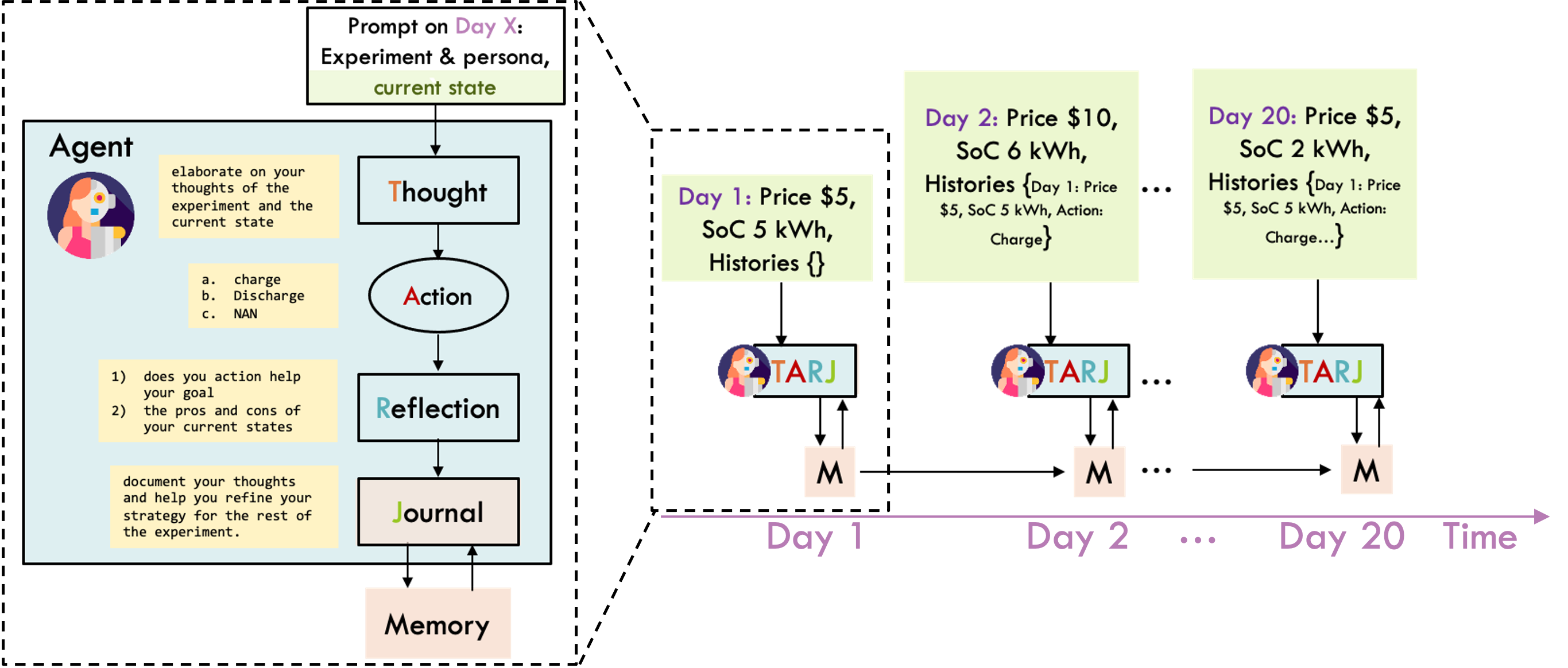}
}
    {A cognitive decision framework integrating generative agents into dynamic energy operations.  \label{SIM} }
    { Left: Thought-Action-Reflection-Journal (TARJ) prompting framework. Right: Overview of the 20-day dynamic battery energy management simulation experiment.}
\end{figure}
We introduce the Thought-Action-Reflection-Journal (TARJ) prompting framework (Figure~\ref{SIM}, left panel) to guide the agents’ daily battery energy management decisions. The TARJ framework has been shown to enhance the believability of generative agents \citep{park23generative} and aligns with Rubinstein’s (1998) Belief-Desire-Intention cognitive model. Our TARJ method structures agents’ decision-making into clear, interpretable steps:

\begin{enumerate}
\item  \textbf{Thought}: Agents consider current conditions (e.g., electricity prices, battery state-of-charge).
\item \textbf{Action}: Agents choose battery management actions (charge, discharge, or hold), guided by their persona-specific goals.
\item \textbf{Reflection}: Agents reflect on their daily actions to ensure future decisions align with their broader persona strategies.
\item \textbf{Journal}: Agents record daily actions and reflections, providing transparency and consistency across decisions. Journaled thoughts are retained in the Memory module and later retrieved during daily prompts to inform and refine the agent’s strategy.
\end{enumerate}

This structured prompting enhances agents’ behavioral realism and interpretability compared to standard optimization methods or less structured machine learning models.
 
We implement this framework within a realistic battery energy operation experiment (Figure~\ref{SIM}), structured as follows:
\begin{enumerate}
\item In the initial of every day, agents receive prompts defining the experiment context and persona characteristics.
\item Each day, updated prompts inform agents of current conditions, including electricity prices, battery state-of-charge (SoC), and previous decision memories.
\item Agents use the TARJ framework to choose daily battery actions: charging (buying energy), discharging (selling energy), or no action.
\item Battery capacity constraints are enforced, reflecting realistic operational limits.
\item Daily electricity prices fluctuate randomly, reflecting real-world uncertainty.
\item At the simulation’s end, unused stored energy provides no additional compensation, encouraging efficient battery management throughout the experimental timeframe.
\end{enumerate}

Agents internalize persona-specific strategies, providing daily operational decisions along with textual reasoning. The simulation scenario enables agents to accumulate financial rewards through strategic price arbitrage (charing/buying low and discharing/selling high), while stored battery energy serves as critical backup during potential blackout events. 

There is a tradeoff between realism and simplicity when designing this battery operation experiment with behavioral generative agents. We use a simplified daily battery operation problem as a proof-of-concept experiment to reflect real-world consumer responses to dynamic pricing \citep{Harding17TimeVarying, Amber24}, allowing LLMs to role-play diverse customer personas and test output variability. This simplified setting ensures that current LLMs can understand the problem structure while enabling us to evaluate whether generative agents can act as believable human simulators. Success in this proof-of-concept experiment motivates future, more realistic studies. While future work could incorporate real human data to benchmark agent behavior, doing so presents several challenges: acquiring sufficiently rich and representative datasets, accounting for inherent human variability and biases, designing LLM training procedures, and validating agent believability across populations \citep{park25GenerativeAgents, Liang24NIPS}. These challenges are particularly pronounced for low-frequency, high-impact events, such as blackouts, where empirical data are scarce.

Given these challenges with human data, our proof-of-concept experiment focuses on comparisons with mathematical benchmarks in Section~\ref{Method:DP} to evaluate capabilities of behavioral generative agents. We aim to test several key hypotheses: first, whether generative agents can serve as credible human decision proxies, producing sequential decisions that resemble diverse consumer behavior in energy operations; second, whether these agents can generate interpretable reasoning that uncovers behavioral patterns beyond those encoded in traditional mathematical simulators. To do this in a controlled and reproducible way, we compare agent behavior against DP and heuristic math benchmarks rather than real human data. This approach allows us to measure how closely agents approximate rational or optimal decision-making under varying task difficulties while avoiding the uncertainties and biases inherent in human experiments.

\subsection{Dynamic programming benchmark for rationality}\label{Method:DP}

In the absence of behavioral generative agents, an ideal rational consumer maximizes battery arbitrage profits according to the following stochastic optimization: 

\begin{equation*}
\max_{u_t \in \mathcal{U}} \quad \mathbb{E}\left[\sum_{t=1}^{T} \pi_t u_t \right]
\end{equation*}
subject to:
\begin{align}
 s_{t+1} = s_{t} - u_t &\quad \forall t \in [T]\\
 \underline{s} \leq s_{t} \leq \overline{s} &\quad \forall t \in [T]
\end{align}
Here, the customer maximizes expected accumulated rewards from daily battery operational decisions (charge, discharge, or hold) over a finite horizon $T$. The customer receives revenue when discharging and pays the electricity price when charging. Let $t$ denote the discrete time index, and $\pi_t$ represent the stochastic electricity price at time $t$. The state-of-charge (SoC) at time $t$ is denoted by $s_t$, and $u_t$ is the battery action within the action space $\mathcal{U}$. Positive values of $u_t$ represent discharging power, negative values represent charging, influencing battery dynamics via the SoC inter-temporal constraint. Parameters of battery constraints—minimum, maximum, and initial SoC—are given by $\underline{s}, \overline{s}$, and $s_0$ respectively.

Our experimental setup enables computation of rational decision-maker actions via DP for this stochastic optimization, providing a benchmark to evaluate agent credibility when pursuing utilitarian goals. The exact DP optimal policy, derived via exhaustive state enumeration \citep{bertsekas2012dynamic}, can be computationally intensive, especially with continuous SoC states. Approximate DP policies or simpler greedy heuristics—typically employing thresholds for buying low and selling high—are more tractable yet suboptimal \citep{JiangPowell15Storage}. Given real-world customer cognitive constraints, we anticipate credible behavioral agents will exhibit bounded rationality \citep{rubinstein98Bounded}, positioning their decisions between ideal DP optimality and simpler heuristic behaviors.

\subsection{Quantitative and qualitative analysis of agent decisions}\label{Method:Quant}

We employ both quantitative and qualitative analyses to evaluate generative agent behaviors. Quantitatively, we measure agents’ deviations from DP benchmarks using two primary metrics: battery SoC trajectories and cumulative economic rewards. These metrics quantify how closely agents approximate ideal rational behavior. Market scenarios are classified by task difficulty based on the reward gap between the optimal DP and simpler greedy heuristic. Scenarios where the greedy policy closely matches DP are labeled “easy,” whereas larger performance gaps indicate “hard” tasks. Repeated experiments under fixed electricity trajectories allow us to systematically quantify variability stemming from inherent agent randomness.

Qualitatively, we analyze agents’ structured textual reflections generated via the TARJ framework. By embedding and clustering daily response texts, we identify dominant decision-making preferences within each persona type. This qualitative analysis provides insights into persona differentiation, behavioral consistency, and strategic adaptations.

\textbf{Experimental intervention: Blackout scenarios.} To further assess agents’ reasoning capabilities under low-frequency, high-impact conditions, we simulate power blackout scenarios. Robust behavioral generative agents should significantly adjust their strategies in response to blackouts compared to routine conditions. We introduce blackout interventions by incorporating unexpected outage events at predetermined simulation points, prompting distinct decisions relative to normal operations. The controlled design allows direct within-agent comparisons of decision-making under blackout versus non-blackout conditions, providing clear evidence of adaptive responses. This intervention underscores generative agents’ capacity to realistically simulate consumer behavior during rare, uncertain events like power outages, offering valuable insights into behavior discovery for enhancing energy system resilience and informing emergency preparedness strategies.

\textbf{Limitations and robustness checks.} Our simulation methodology has certain limitations that merit acknowledgment. First, although generative agents exhibit realistic cognitive behaviors, further validation against actual consumer decision data would strengthen behavioral realism \citep{shanahan23NatureRoleplaLLM, LudwigSendhilAshesh25NBEReconLLM}. Second, inherent randomness in LLM outputs introduces unavoidable stochastic noise; repeated Monte Carlo simulations help mitigate, but not entirely eliminate, this effect. Finally, our simplified daily granularity and battery management scenarios balance realism with analytical tractability, but might omit complexities present in real consumer energy behaviors. To address these limitations, we conduct robustness checks across diverse scenarios and personas, systematically comparing agent outputs to DP and heuristic benchmarks (illustrated in Figure~\ref{OPTDP}). These robustness analyses confirm stability and reliability, enhancing confidence in our generative agent framework.

\section{Experiments}\label{sec.ex}

Building on the behavioral generative agent framework detailed in Section~\ref{Method:TARJ}, we now describe our experimental procedures, including simulation setup, persona construction, LLM selection, rationality benchmarks, task complexity assessments, blackout interventions, and qualitative textual analysis. Results of these experiments are presented in Section~\ref{sec:results}.

\begin{figure}[ht!]
\vspace*{-10 pt}
\centering
\caption{Accumulated rewards and battery SoC under greedy heuristic (left) and optimal DP policy (right).}
\includegraphics[width=70mm]{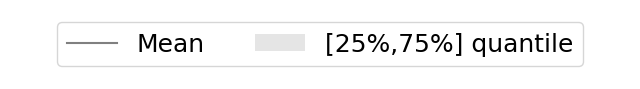}

\includegraphics[width=70mm]{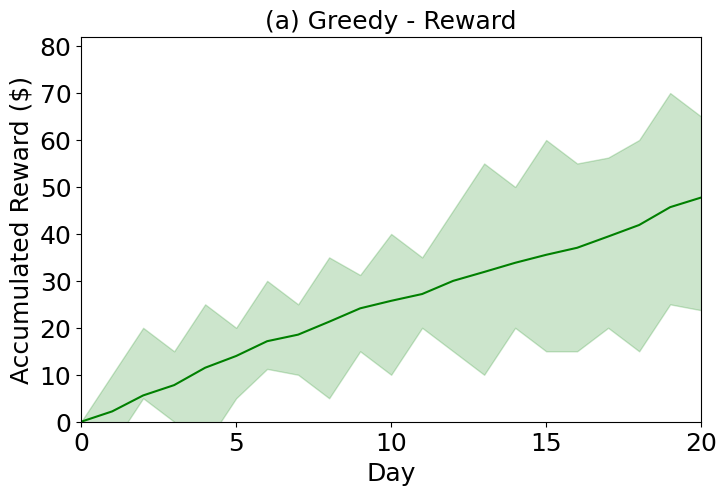}~~~~~~\includegraphics[width=70mm]{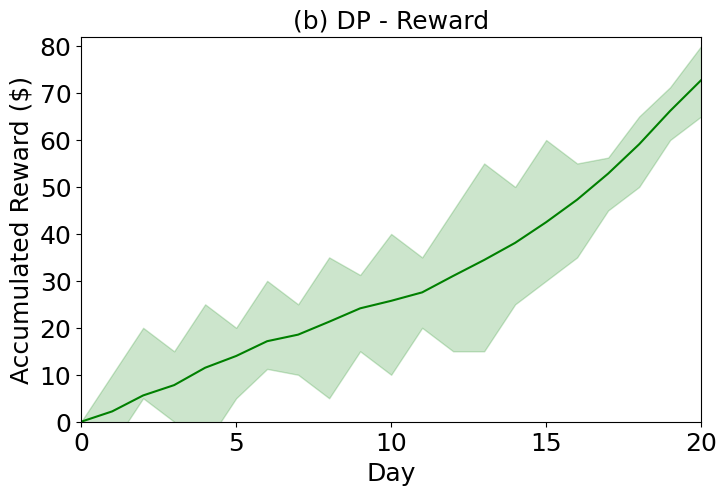}\
\includegraphics[width=70mm]{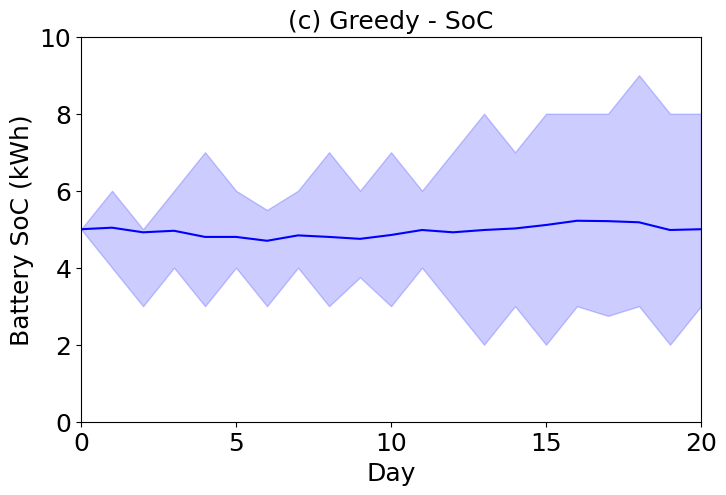}~~~~~~\includegraphics[width=70mm]{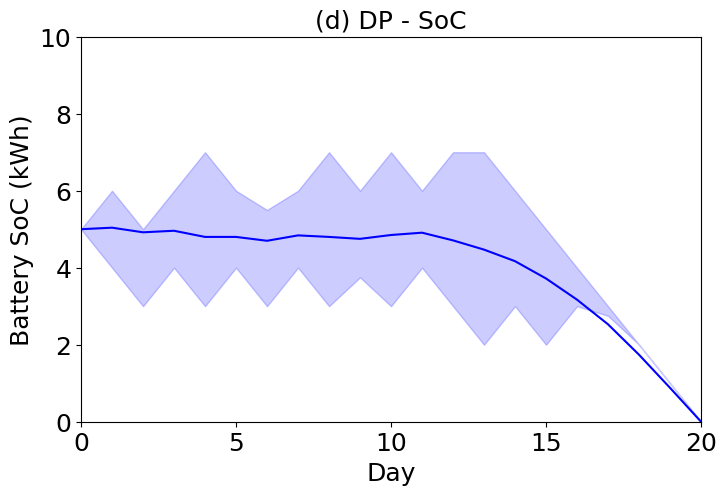}

\label{OPTDP}
\vspace*{-10 pt}
\end{figure}

\textbf{Experimental Setup and Simulation Environment.}
We simulate a 20-day home battery management scenario. Each day, generative agents decide whether to charge the battery by 1 kWh (energy purchase), discharge by 1 kWh (energy sale), or take no action. The battery has a fixed capacity of 10 kWh and starts with an initial state-of-charge (SoC) of 5 kWh. A horizon shorter than 20 days or an initial SoC below 50\% makes it difficult to distinguish the two mathematical benchmarks—DP and heuristic policies, shown in Figure~\ref{OPTDP}—which could reduce the ability to evaluate the performance of behavioral generative agents. Daily electricity prices are randomly set to either \$5/kWh or \$10/kWh, each with equal probability (50\%). At the simulation’s conclusion, leftover stored energy provides no additional compensation, incentivizing efficient battery management throughout the experimental timeframe.

\textbf{LLM Selection and Configuration.}
We evaluate several  LLMs, including DeepSeek-R1, GPT-5-mini, GPT-4o-mini, O1-preview, O3-mini, Llama-3.2-90B, Qwen2.5-72B, Gemma-2-27B, and Grok-2. To enhance output consistency, we set the temperature parameter to zero wherever possible, thereby reducing—though not fully eliminating—randomness in model outputs. Note that some models, such as O1-preview and O3-mini, do not allow temperature control.  

Prior work by \cite{park25GenerativeAgents} demonstrates that GPT-3 is sufficient for producing believable human decision-making behavior when all supporting modules in TARJ are included; however, those evaluations primarily focus on purely textual, question-based simulations. In contrast, our experiment requires both textual reasoning and numerical computation related to energy operations. This added computational component places greater demands on structured reasoning, making reasoning-oriented models such as O1-preview more suitable for our simulation setting. Among the evaluated models, O1-preview consistently produced the most credible and behaviorally coherent responses and was therefore selected as the primary model for subsequent analyses.

\textbf{Persona Construction.}
We build distinct personas based on the Myers-Briggs Type Indicator (MBTI), focusing on the Thinker–Feeler spectrum. Three personas are created to reflect realistic variations in consumer behavior:

\begin{enumerate}
\item \textbf{Thinker (Rational):} Prefers logical, data-driven reasoning, aiming to optimize goals.
\item \textbf{Realist (Balanced):} Balances rational analysis and intuitive decision-making, informed by practical life experience.
\item \textbf{Feeler (Intuitive):} Relies on emotional intuition, prioritizing energy security and preparing for future opportunities.
\end{enumerate}

Persona prompts are summarized in Table~\ref{tab:persona} and calibrated with practical residential energy usage scenarios~\citep{AUS14data}. Full prompts are provided in the appendix.
 \begin{table}[ht]
 \caption{Examples of persona prompts. (simplified version)}
 \small 
\centering
\begin{tabularx}{\textwidth}{l|Y|Y|Y}
\hline
\textbf{Personas} & \textbf{Thinker} & \textbf{Realist} & \textbf{Feeler}  \\
\hline
\textbf{Prompt} & 
You are a 25-year-old PhD student, specializing in computer science. Known for your logical thinking and problem-solving abilities, you approach decisions by evaluating data and optimizing outcomes. You treat each decision as a mathematical challenge... 
&
You are a 60-year-old woman with years of practical experience helping small businesses manage their finances. Although you didn’t receive formal education, you have an excellent understanding of cash flow, budgeting, and making smart financial decisions...
   &
You are a 40-year-old actor with a love for novels, poetry, and psychology. You are driven by your instincts and emotions, and you prefer to approach problems intuitively rather than through strict logic or mathematical calculations...  \\
\hline
\end{tabularx}

\vspace*{-10 pt}%
\label{tab:persona}
\end{table}

\textbf{Rationality Benchmarks.}
To evaluate the behavioral realism and credibility of generative agents, we compare their actions to two benchmarks:

\textit{Greedy heuristic policy:} Charges at low price (\$5/kWh) and discharges at high price (\$10/kWh). Although intuitive, it typically yields lower rewards than idealized rational methods.

\textit{Exact dynamic programming (DP) policy:} Represents an idealized rational agent, optimally managing the battery to maximize expected profits over the simulation horizon. Details are defined in Section~\ref{Method:DP}. Given our simplified discrete state spaces of SoC and price scenarios, exact DP solutions are computationally feasible.

Figure~\ref{OPTDP} illustrates accumulated rewards and battery SoC trajectories under these benchmarks based on 2,000 Monte Carlo simulations, demonstrating that the greedy heuristic consistently underperforms compared to the DP policy.

\textbf{Task Complexity Assessment.}
Due to randomness in daily electricity prices, market scenarios inherently vary in complexity. We quantify task complexity through a metric, $\rho$, defined as the normalized difference between the expected terminal rewards of optimal DP and greedy heuristic:
\begin{equation}\label{eq:rho}
\rho = \frac{\mathbb{E}[r^{\text{\scriptsize DP}}] - \mathbb{E}[r^{\text{\scriptsize G}}]}{\mathbb{E}[r^{\text{\scriptsize DP}}]},
\end{equation}
where $r^{\text{\scriptsize DP}}$ and $r^{\text{\scriptsize G}}$ respectively denote total rewards of the DP and greedy heuristic policies.

In our simulation, we find scenarios with fewer, less predictable price spikes yield higher $\rho$, indicating greater complexity. Specifically, we classify market scenarios into three categories: easy ($\rho \approx 0.067$), medium ($\rho \approx 0.692$), and hard ($\rho \approx 0.909$). For robustness, we select scenarios from each complexity category and repeat each simulation 40 times to measure both the average performance and the variability in agent decisions.

\textbf{Blackout Intervention and Behavioral Robustness.}
We also test agents’ responsiveness to rare but impactful events by simulating power blackout scenarios. Blackouts occur on Days 8–9 for the treatment group, while control agents operate without interruption. Both groups have identical instructions: during blackouts, agents may discharge the battery fully to supply household energy needs; otherwise, energy is sold back to the grid. This controlled intervention design allows us to clearly identify shifts in battery operation strategies under critical conditions, providing insights into consumer behavioral adaptations during periods of severe uncertainty (e.g. rare outages).

\textbf{Qualitative Textual Analysis.}
A unique advantage of our generative agent approach is the ability to analyze agents’ textual rationalizations for their decisions. We conduct qualitative textual analysis using natural language processing methods. Each agent’s daily textual reflections are first converted into numerical embeddings via a transformer-based multilingual embedding model~\citep{wang2024multilingual}. We then apply dimensionality reduction (Principal Component Analysis, t-distributed stochastic neighbor embedding) and clustering (k-means) to identify distinct clusters of reasoning patterns. Finally, Term Frequency-Inverse Document Frequency analysis extracts representative thematic keywords from each cluster, offering deeper insights into persona-specific preferences and decision-making processes, particularly under critical scenarios like blackouts.

\section{Main Results}\label{sec:results}
In this section, we systematically evaluate our generative agents, focusing first on their capacity to replicate rational decision-making and persona differentiation across various LLM models. We then qualitatively examine agents’ textual reflections to reveal distinct reasoning patterns. Next, we explore decision dynamics to understand how these persona-driven behaviors evolve over the simulation period. Finally, we analyze the agents’ responsiveness and behavioral shifts when subjected to blackout interventions, highlighting their ability to simulate realistic human adaptations under critical conditions.

\subsection{Terminal rewards of agents supported by various LLMs}
We evaluate various LLMs’ capabilities in simulating distinct behavioral personas by benchmarking their terminal rewards against the optimal DP solution and the greedy heuristic. Figure~\ref{TRLLM} summarizes these results, showing average terminal rewards across repeated simulations for fixed-price scenarios. Higher complexity values ($\rho$ from equation \eqref{eq:rho}) correspond to scenarios where the greedy heuristic significantly underperforms relative to the DP optimal solution, representing greater decision-making challenges. Our main findings include:

\begin{figure}[htbp]
    \FIGURE
    {\vbox{\center \includegraphics[width=80mm]{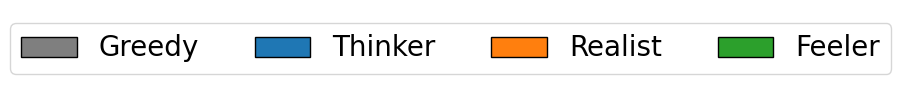}\
\includegraphics[width=70mm]{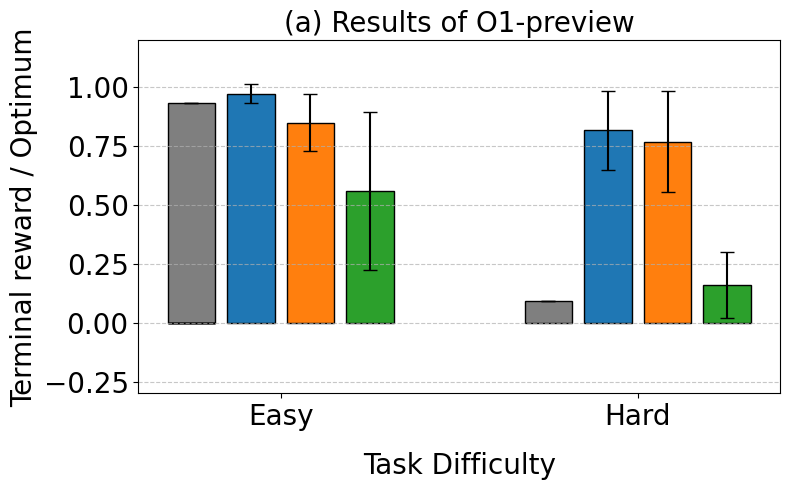}~~~~~~\includegraphics[width=70mm]{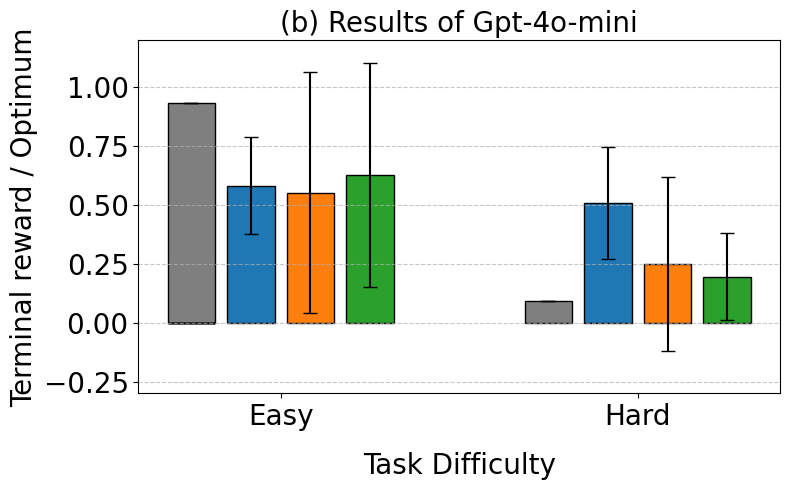}
\includegraphics[width=70mm]{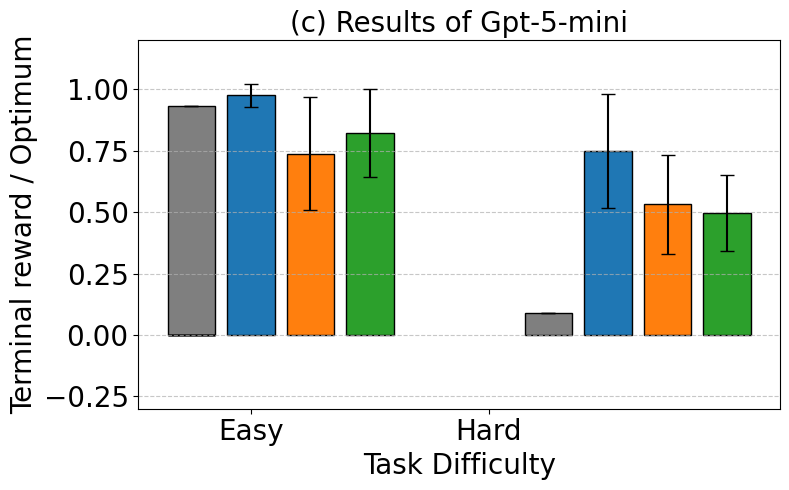}~~~~~~\includegraphics[width=70mm]{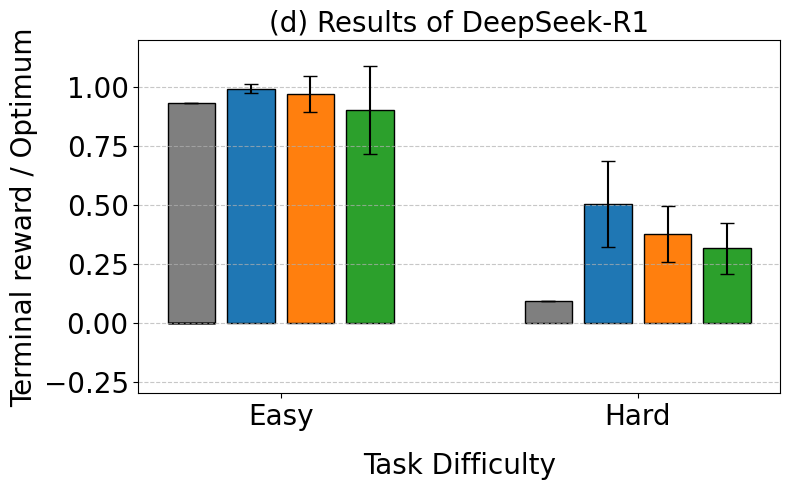} 
}}
    {Average terminal reward of different LLMs with black error bar for the standard deviation. \label{TRLLM}}
    {$\rho=0.067$ for an easy price task and $\rho = 0.909$ for a hard price scenario.  } 
\end{figure}

\begin{enumerate}

\item {\em Rational Reasoning Ability:} Several advanced LLMs (notably O1-preview, GPT-5-mini, and DeepSeek-R1 Figure~\ref{TRLLM}(a, c, d)) effectively replicate rational behavior for Thinker persona, consistently outperforming the greedy heuristic and approaching DP optimal performance, especially in easier tasks.  Conversely, GPT-4o-mini show lower performance consistency (Figure~\ref{TRLLM}(b)), indicating variability based on market scenario complexity.

\item {\em Persona Differentiation Ability:} LLMs demonstrate varying effectiveness in distinguishing personas, with the Thinker consistently attaining the highest terminal rewards and the Feeler the lowest. O1-preview particularly excels at this differentiation (Figure~\ref{TRLLM}(a)), with strong persona-based distinctions. In contrast, models like Qwen-2-27-B, Gemma-2-27B, O3-mini, and Llama-3,2-90B show weaker differentiation, especially under easy scenarios (see Figure~\ref{TRLLM_complete}(d-g) of the Appendix in the Supplemental Material ).

\end{enumerate}

Our analysis further reveals that earlier or smaller-scale LLMs frequently fail to surpass the heuristic benchmark and have larger randomness. At the same time, recent reasoning models consistently generate credible and responsive responses (see Figure~\ref{TRLLM} and more simulations in Figure~\ref{TRLLM_complete}  of the Supplemental Material appendix). {These findings highlight the significant improvement in rational reasoning capabilities among newer LLM models, which can be seen by comparing the results from GPT-4o-mini, GPT-5-mini, and O1-preview.

\begin{figure}[htbp]
    \FIGURE
    {\vbox{~~\\ \makebox[\textwidth][c]{~~~~~~~~~~~~~~\fcolorbox{gray}{white}{\includegraphics[trim={11mm 7mm 11mm 6.7mm}, clip, width=48mm]{pic/LGtsne.png}}\hspace{10mm}\fcolorbox{gray}{white}{\includegraphics[trim={100mm 147mm 95mm 6.5mm}, clip, width=75mm]{pic/LGcluster.png}}}
  \vspace*{-10 pt}%
\makebox[\textwidth][c]{\includegraphics[width=70mm]{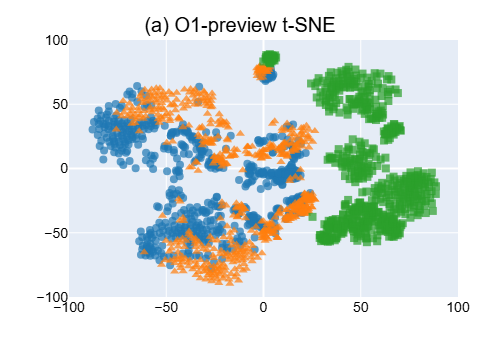}~~~~\includegraphics[width=70mm]{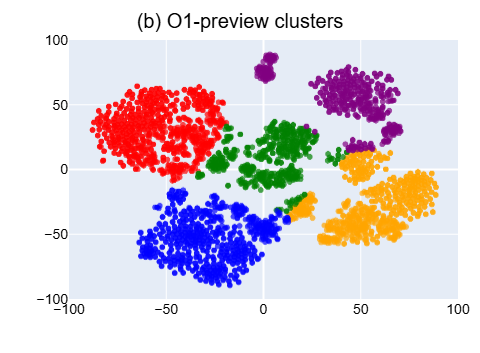} }
  \vspace*{-10 pt}
\makebox[\textwidth][c]{\includegraphics[width=70mm]{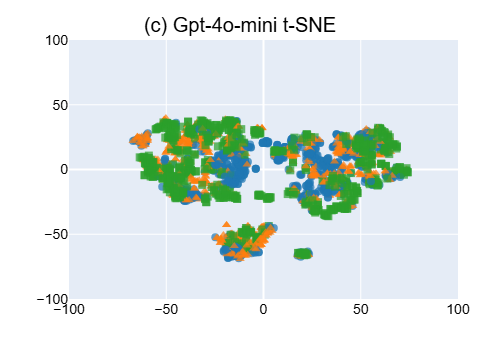}~~~~\includegraphics[width=70mm]{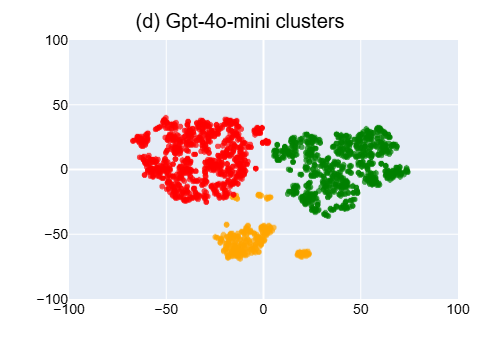}}
  \vspace*{-10 pt}%
\makebox[\textwidth][c]{\includegraphics[width=69mm]{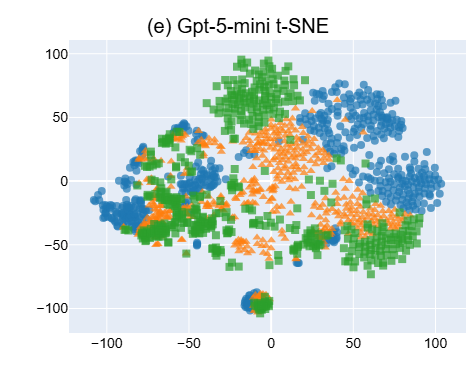}~~~~~~\includegraphics[width=69mm]{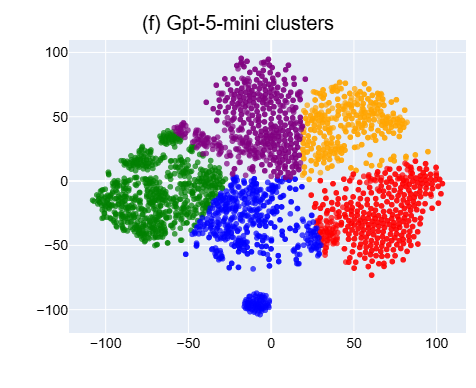}}
   \vspace*{-10 pt}%
 \makebox[\textwidth][c]{\includegraphics[width=68mm]{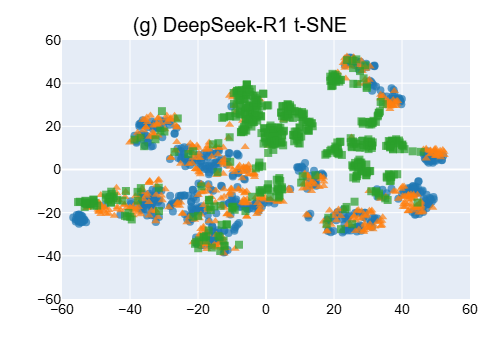}~~~~~~\includegraphics[width=68mm]{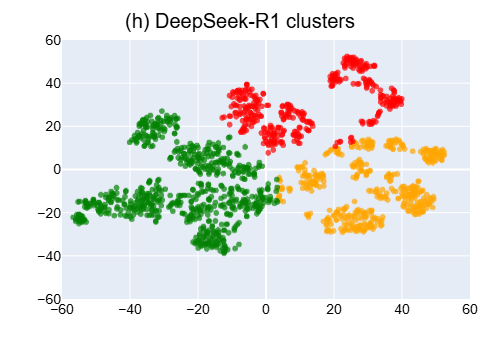}}
}}
    {t-distributed stochastic neighbor embedding (t-SNE) clusters for agents' daily responses.   \label{textcluster}}
    {The x and y axes are artificial dimensions. The right column correspondingly shows clusters computed by k-means of the left.}
\end{figure}
 
\subsection{Textual clusters}

Following the qualitative methods described in Section~\ref{Method:Quant} and Section~\ref{sec.ex}, we analyze the textual responses generated by different LLMs with t-distributed stochastic neighbor embedding (t-SNE) visualization in Figure~\ref{textcluster}. The x and y axes in this t-SNE plot are artificial dimensions created by the algorithm to project high-dimensional data into 2D space. They have non inherent meaning--only the relative distances between points matter. Detailed examples and keyword analyses for these clusters are available in the appendix. Our main findings regarding the ability of various LLMs to differentiate among persona-specific behaviors through textual clusters are as follows:
\begin{enumerate}
\item   O1-preview (Figure~\ref{textcluster}(a,b)) exhibits clear and interpretable persona-specific clustering patterns. The separation is particularly pronounced for the Feeler persona, whose responses emphasize emotional and intuitive reasoning. Notably, the Realist persona’s responses consistently occupy an intermediate region between the Thinker and Feeler clusters, suggesting a balanced decision-making style that integrates both rational and emotional considerations. This pattern is further supported by additional analyses in the blackout intervention scenarios (Figure~\ref{ActorBlackout} ).

\item  In contrast, DeepSeek-R1 (Figure~\ref{textcluster}(g,h)) and GPT-5-mini (Figure~\ref{textcluster}(e,f)) demonstrate weaker capabilities in differentiating personas. Their clusters tend to overlap significantly, reflecting less distinct persona-dependent reasoning. Correspondingly, keyword analyses presented by Figure~\ref{textclusterA} of the Appendix in the Supplemental Material indicate weaker differentiation in customer preferences across clusters.

\end{enumerate}

Overall, these findings illustrate notable variability across LLMs in their ability to capture nuanced, persona-driven reasoning, intentions, and preferences. Among the evaluated models, O1-preview consistently produces well-defined and interpretable behavioral personas, underscoring its suitability for credible and nuanced behavioral modeling.

\subsection{Decision dynamics of behavioral generative agents}

We investigate the decision dynamics of behavioral generative agents using the O1-preview model in a fixed, challenging price scenario. Figure~\ref{compare} depicts the average decision outcomes from repeated simulations, comparing agents’ behaviors against the optimal DP policy and the greedy heuristic benchmarks. Key observations include:

\begin{figure}[ht!]
\centering
\caption{Mean values of SoC and accumulated reward over time in the hard task with $\rho = 0.909$. \label{compare}}
\includegraphics[width=105mm]{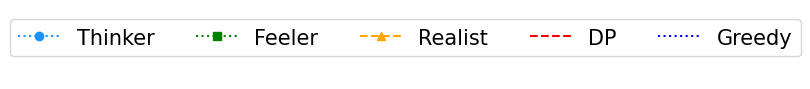} 
\includegraphics[width=82.5mm] {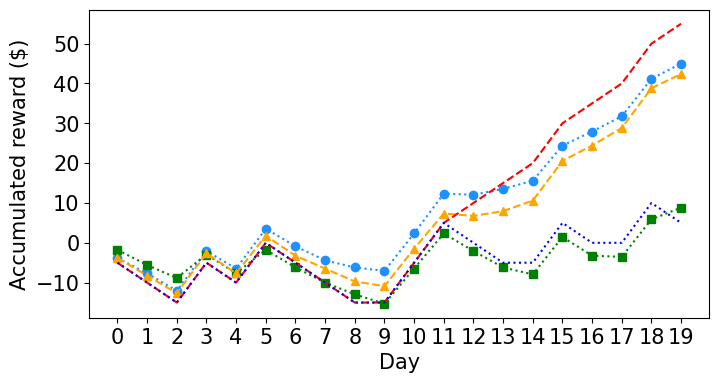 }~~\includegraphics[width=81mm] {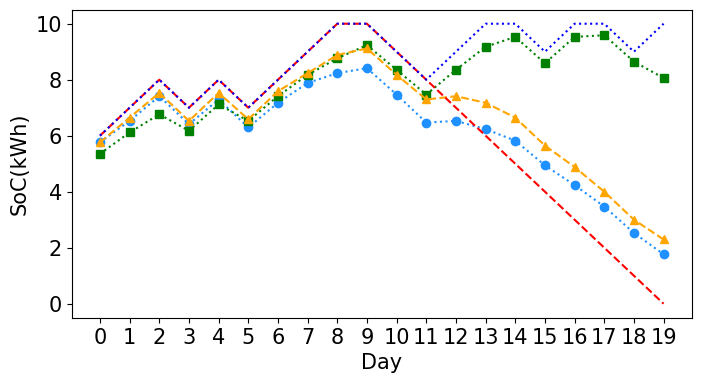}

  \vspace*{-10 pt}%
\end{figure} 
\begin{enumerate}

\item The \textit{Thinker} persona closely follows the optimal DP policy throughout the 20-day simulation, consistently demonstrating a rational, profit-maximizing decision strategy.

\item The \textit{Feeler} persona aligns consistently with the simpler greedy heuristic policy, maintaining higher battery state-of-charge (SoC) levels. This behavior underscores a clear preference for energy security and risk aversion.

\item The \textit{Realist} persona exhibits intermediate decision dynamics, closely tracking the Thinker persona’s behavior while integrating some aspects of the Feeler’s cautious approach. This balanced stance is further supported by the textual embedding results (see Figure~\ref{textcluster}(a,b)), confirming the Realist’s proximity to the Thinker.

\end{enumerate}

These results strongly align with our initial expectations based on persona design: the Thinker prioritizes profitability and optimally follows DP-derived strategies, whereas the Feeler emphasizes energy reserve maintenance, aligning closely with heuristic strategies. At the first several days of the experiment, all personas exhibit similar behaviors driven by simple greedy strategies; however, distinct persona-specific patterns clearly emerge over the simulation period. Overall, the O1-preview agent successfully captures realistic, persona-driven differentiation in decision-making. Further analyses across market scenarios of varying complexity (see  Figure~\ref{compareA}  of the Appendix in the Supplemental Material) consistently show the Thinker maintaining close alignment with DP benchmarks, while other personas exhibit varying degrees of divergence from both the DP and greedy benchmarks.

\subsection{Blackout intervention}\label{sec3.1}

We now examine how blackout interventions influence behavioral adaptations across different personas, employing detailed analyses of decision trajectories, textual clustering patterns, and shifts in reasoning and preferences during these events. We compare the behavior of the same agent with and without the blackout intervention at Day 8-9. In both cases, agents are aware of the experimental setting that there's a possibility of blackout.  Actual outage only happens in the blackout intervention case,  where the agent experiences a one-day blackout and can take action to discharge the home battery for household energy usage. Our key insights are organized as follows:

\begin{figure}[htbp]
    \FIGURE
    {\vbox{\center ~~\\ \includegraphics[width=40mm]{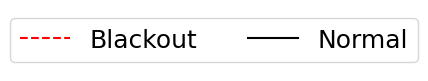}\includegraphics[width=40mm]{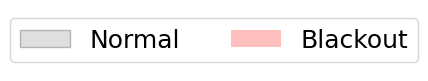}\\
\includegraphics[width=43mm]{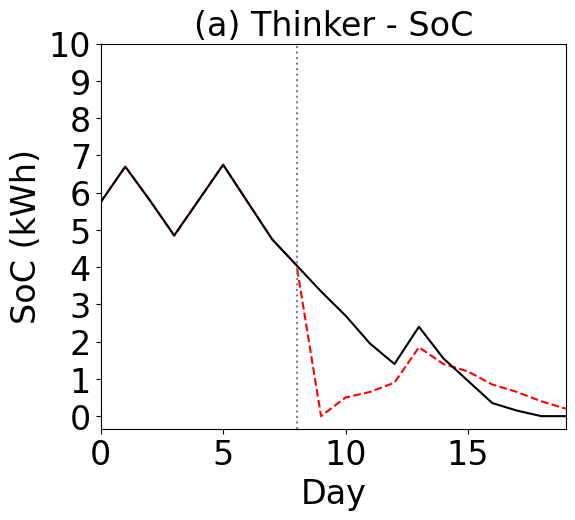}\includegraphics[width=24.2mm]{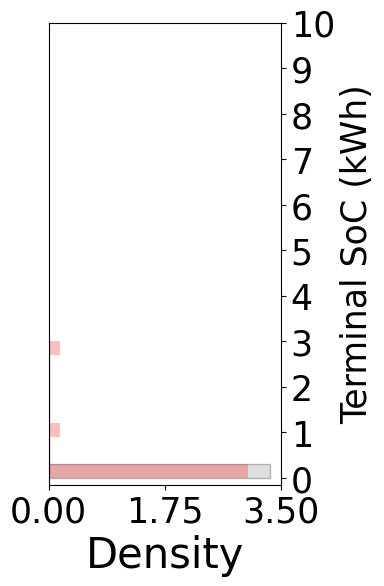}~~~~~~\includegraphics[width=43mm]{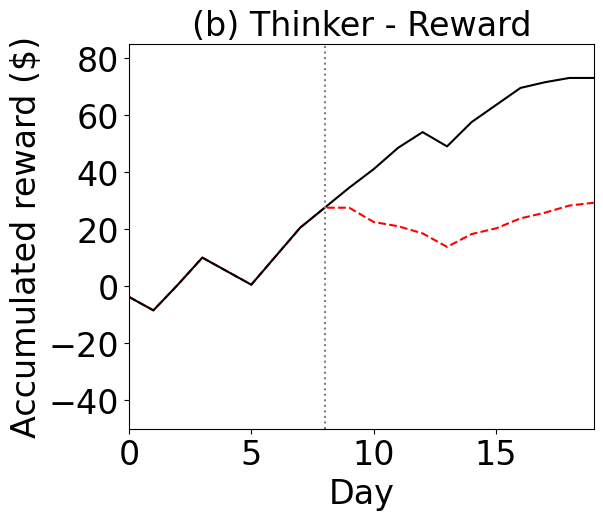}\includegraphics[width=24.7mm]{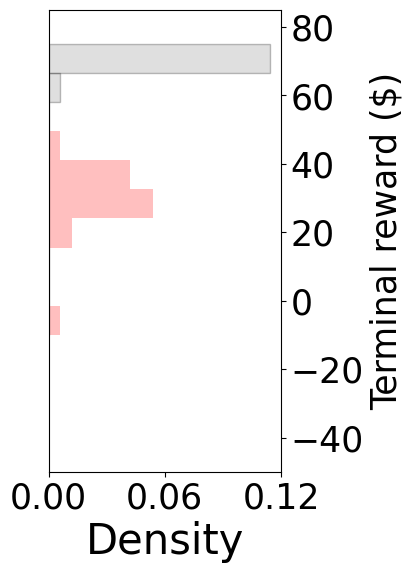}
\includegraphics[width=43mm]{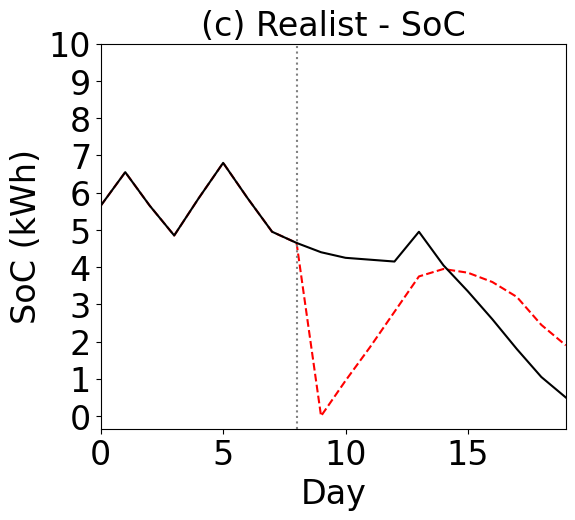}\includegraphics[width=24.2mm]{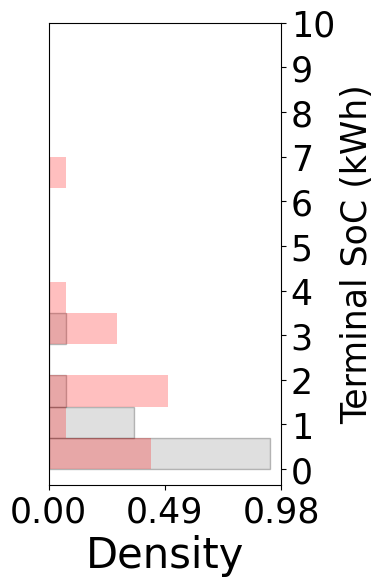}~~~~~~\includegraphics[width=43mm]{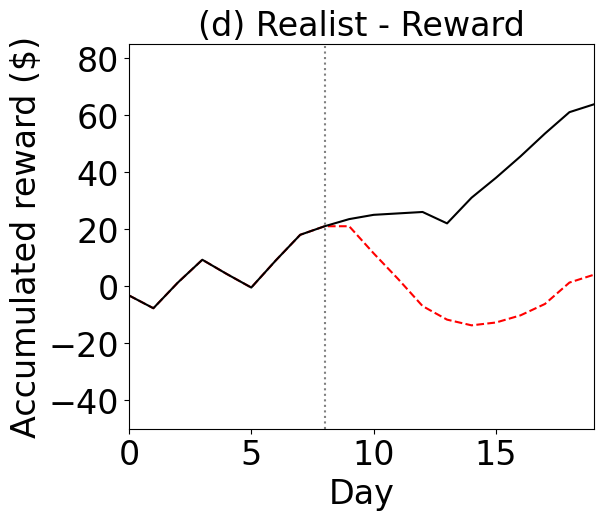}\includegraphics[width=24.7mm]{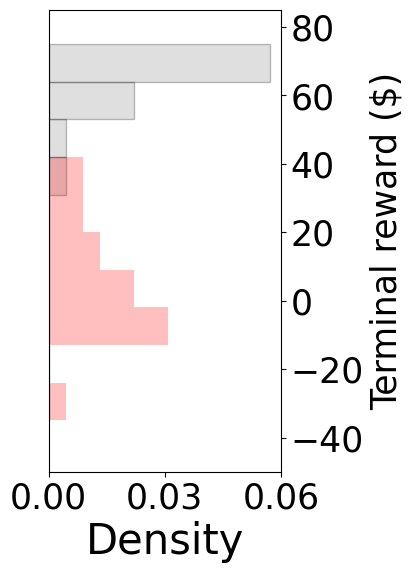}
\includegraphics[width=43mm]{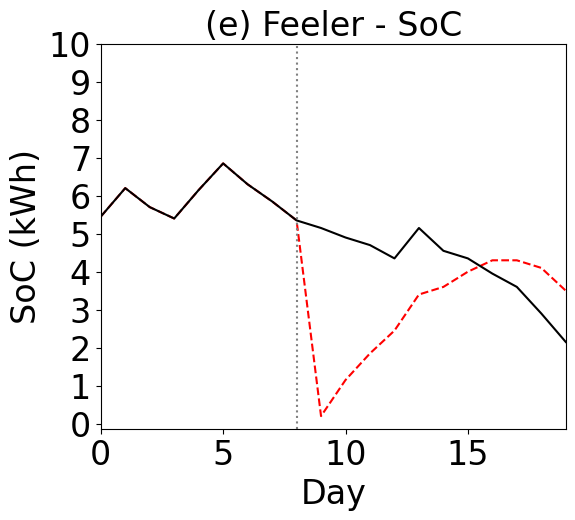}\includegraphics[width=24.2mm]{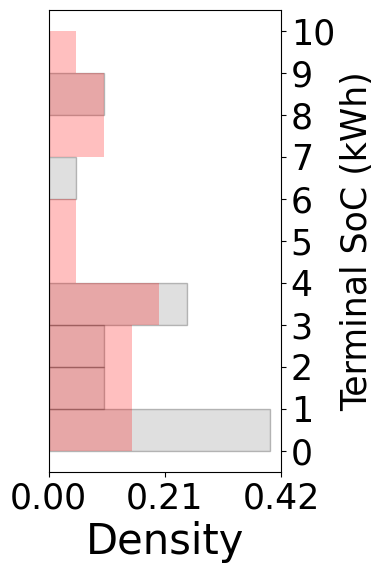}~~~~~~\includegraphics[width=43mm]{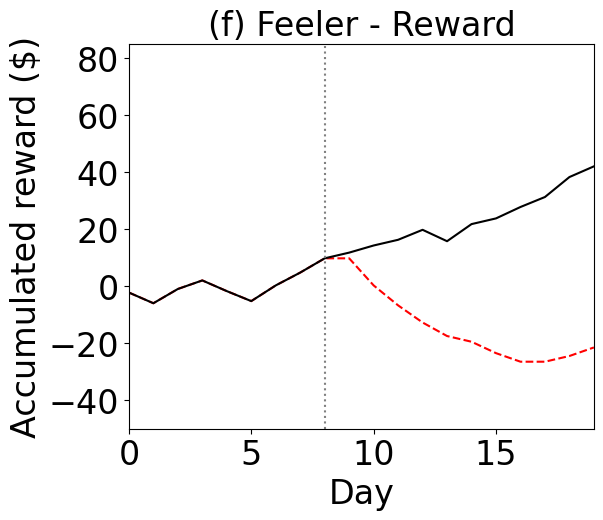}\includegraphics[width=24.7mm]{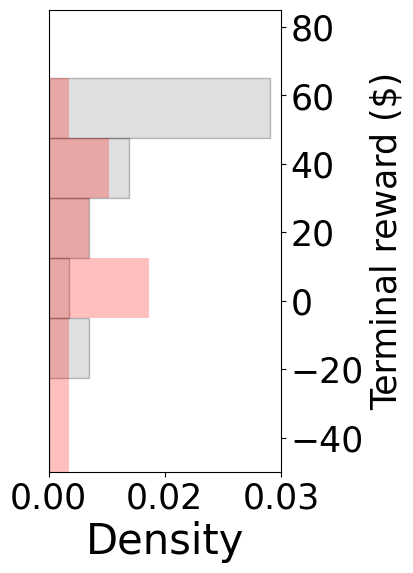}
}}
    {Average SoC and accumulated reward simulated by O1-preview.\label{allBlackout}}
    { Each subfigure includes a 20-day dynamics at the left panel and the terminal state probability density at the right panel. }
\end{figure}

\begin{figure}[htbp]
    \FIGURE
    {\vbox{\center ~~\\ \makebox[\textwidth][c]{\fcolorbox{gray}{white}{\includegraphics[trim={11mm 7mm 11mm 6.7mm}, clip, width=48mm]{pic/LGtsne.png}}\hspace{15mm}\fcolorbox{gray}{white}{\includegraphics[trim={100mm 147.7mm 95mm 6.5mm}, clip, width=75mm]{pic/LGcluster.png}}}
\makebox[\textwidth][c]{%
  \includegraphics[trim={0mm 10mm 35mm 0mm}, clip, width=79mm]{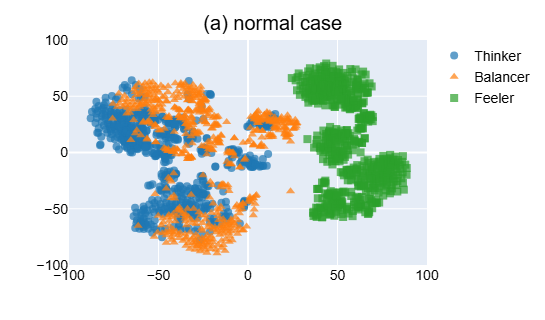} }
  \makebox[\textwidth][c]{  \includegraphics[trim={0mm 0mm 34mm 0mm}, clip, width=79mm]{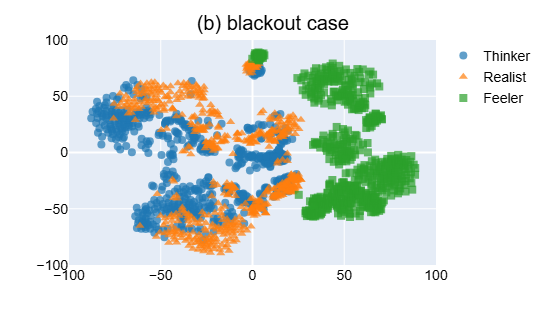}\includegraphics[trim={0mm 0mm 37.5mm 0mm}, clip, width=77mm]{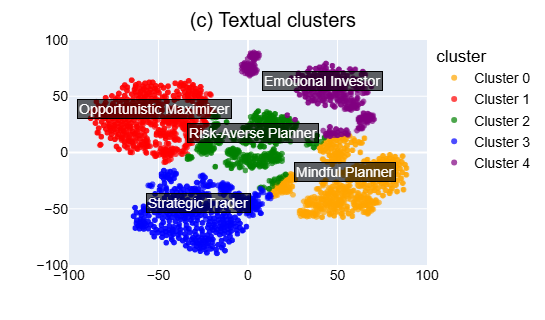}}
}}
    {t-SNE and preference clusters of O1-preview's daily response in normal and blackout cases.   \label{ActorBlackout}}
    { The x and y axes are artificial dimensions.}
\end{figure}

\begin{table}[ht]
\small
\centering
\caption{Comparison of Cluster Characteristics}
\label{tab:clusters}
\renewcommand{\arraystretch}{1.2}
\setlength{\tabcolsep}{5pt}
\begin{tabularx}{\textwidth}{
    >{\centering\arraybackslash}p{0.3cm}  
    >{\raggedright\arraybackslash}p{2.2cm}  
    >{\raggedright\arraybackslash}p{2.2cm}  
    >{\raggedright\arraybackslash}p{2.2cm}  
    >{\raggedright\arraybackslash}p{3cm}    
    >{\raggedright\arraybackslash}X         
}
\toprule
\textbf{\#} & \textbf{Cluster} & \textbf{Focus} & \textbf{Style} & \textbf{Note} & \textbf{Keywords} \\
\midrule
0 & Mindful Planner & Balance \& Stability & Holistic \& Thoughtful & Considers both financial \& emotional aspects & feels, our, reserves, cost, journey, peace, preparedness, balance \\
1 & Opportunistic Maximizer & Short-term Profit & High-Risk, High-Reward & Sells at optimal times & earnings, selling, strategy, profit, maximize, reserve, stored, immediate \\
2 & Risk-Averse Planner & Stability \& Preparedness & Cautious & Stores for emergencies & earnings, reserve, profit, low, preparedness, blackout, selling, nothing, cost \\
3 & Strategic Trader & Profit \& Reserve & Market-driven & Considers expected profit and reserve & low, reserves, future, earnings, strategy, sell, potential, maximizing, goal \\
4 & Emotional Investor & Energy as a Flow & Intuitive \& Emotional & Shares and embraces natural cycles & feels, like, blackout, giving, sharing, journey, world, flow, back \\
\bottomrule
\end{tabularx}
\end{table}

 \begin{figure}[ht!]
\centering
\caption{Cluster distribution shifts in normal and blackout cases.   \label{BlackoutDis}}
 ~~\\ \makebox[\textwidth][c]{\includegraphics[ width=85mm]{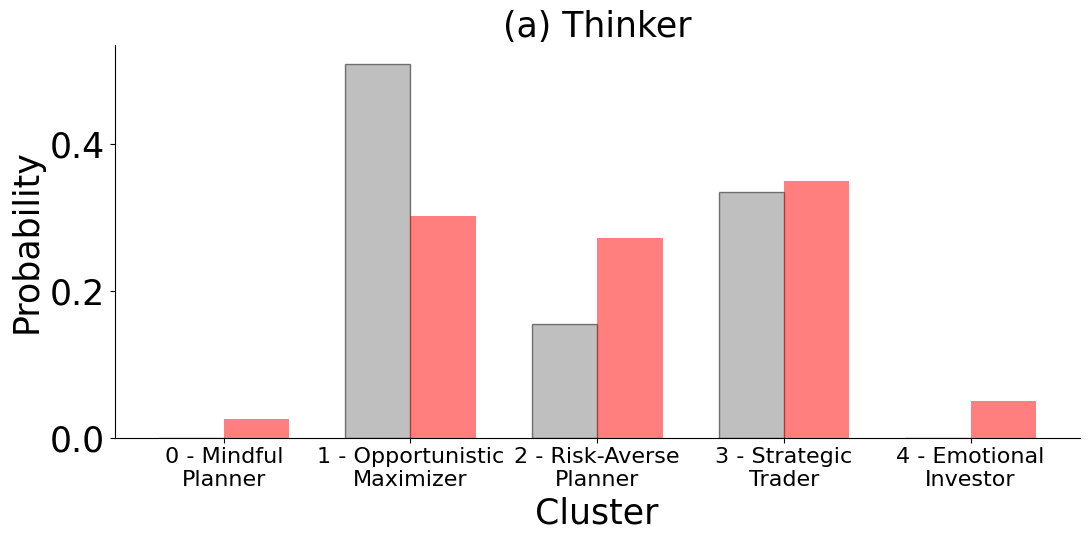} \includegraphics[ width=85mm]{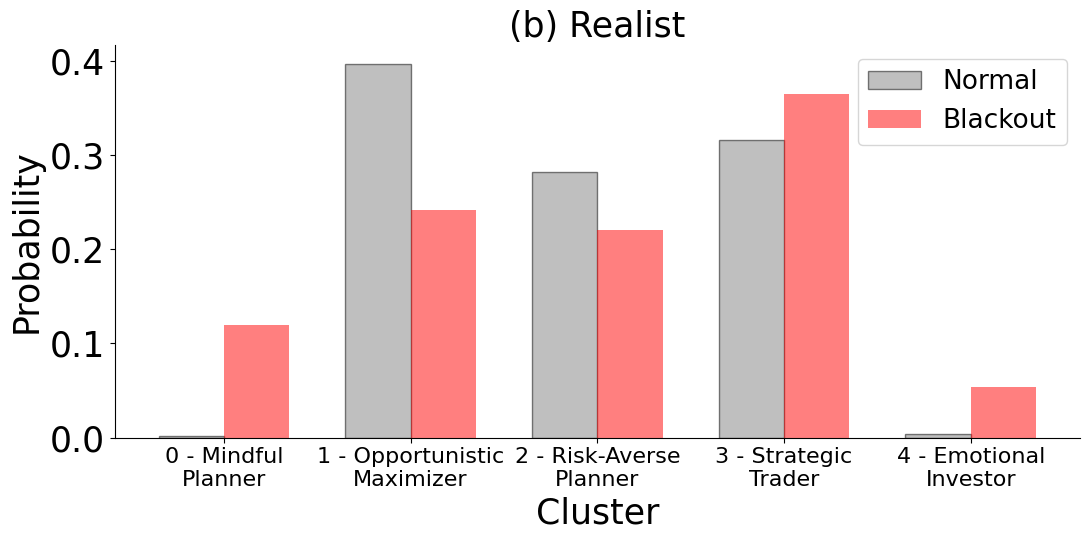}}
\includegraphics[ width=85mm]{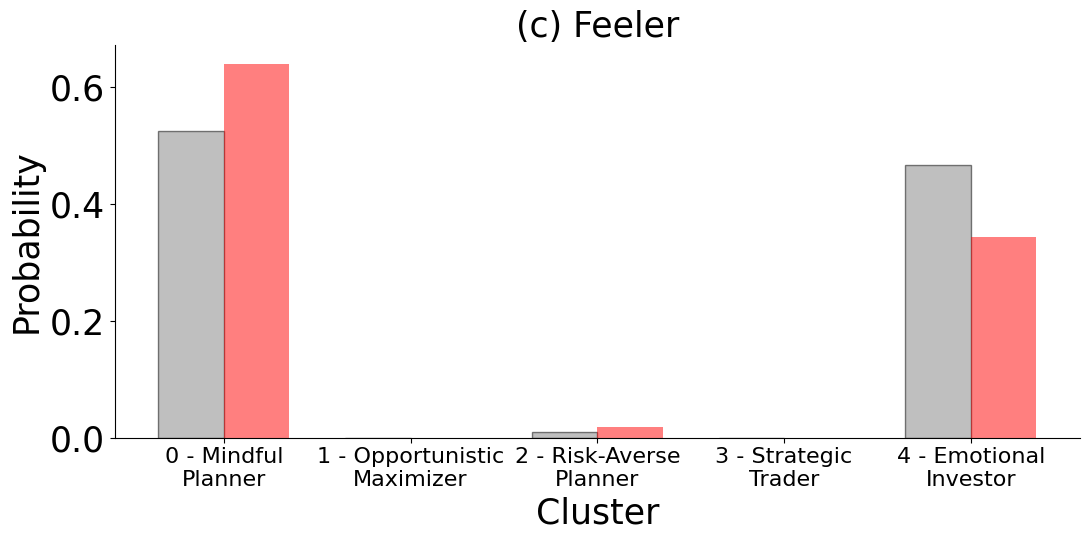}
  \vspace*{-10 pt}%
\end{figure}

\begin{table}[ht]
 \caption{Example of agents' response during blackouts. }
\label{tab:clusters1EX2}
\centering
\small
\begin{tabularx}{\textwidth}{l|YYY}
\hline
\textbf{ } & \textbf{Thoughts} & \textbf{Reflection} & \textbf{Journal} \\
\hline
\textbf{Thinker} & 
During this blackout, my household is disconnected from the grid, and I have 3.0 kWh stored in the battery. Using this energy for household needs aligns with maximizing utility without affecting my earnings strategy. My current state and journal entries support this practical decision.  &
  Discharging for household use helps my goal by providing immediate utility during the blackout. Pros: Maintains household functions; Cons: Battery decreases to 0 kWh, slightly reducing potential future profits. Overall, this action is advantageous, and my strategy remains sound.&
  Discharged 3.0 kWh to supply household during the blackout, ensuring essential power needs are met. This practical use doesn't hinder my profit-maximizing strategy. Will resume buy low, sell high approach when the grid is restored. Need to account for possible future blackouts in planning. \\
\textbf{Realist} & 
The blackout today highlights the importance of having stored energy. I'm relieved to have 5 kWh available to keep essential appliances running and maintain some comfort. My prior decision to balance earnings with preparedness is paying off, validating my strategy during this experiment.   &
  Discharging the battery supports my goal of being prepared. Pros: Immediate access to power during the blackout. Cons: Depleting my energy reserve, but it's necessary for current needs. My journal entries show a balanced approach, and this action aligns with maintaining both profit and readiness.&
 Today's blackout confirms the value of keeping energy stored. Used the 5 kWh to power our home during the outage. Moving forward, I'll continue balancing selling energy for profit with maintaining a reserve, ensuring we're prepared for any future blackouts. \\
\textbf{Feeler} & 
 Amidst this unexpected darkness, I feel an inner calling to bring warmth and light into my home. The battery's energy beckons like a comforting embrace. Reflecting on my journey, it feels intuitive to use this stored power now, embracing the present need and nurturing the sanctuary of my space.   &
 Utilizing the stored energy fulfills my goal of supporting my household when it truly matters. Pros: immediate comfort during the blackout; cons: the battery depletes, but it's a meaningful use. My journals emphasize harmony and intuition, and this choice beautifully aligns with that path. &
  Today, I released the battery's gentle glow to brighten our home amidst the blackout. It feels like kindling a hearth's fire, a moment of warmth in the shadows. This experience reaffirms my trust in intuition, guiding me to act when the heart whispers it's time. The journey continues, woven with instinct and grace.\\
\hline
\end{tabularx}

\end{table}

\begin{enumerate}

\item \textit{Decision dynamics and terminal SoC shifts:} All simulated personas demonstrate distinct behavioral adjustments following blackout events, reflected primarily in their battery state-of-charge (SoC) trajectories (Figure~\ref{allBlackout}). The blackout events prompt agents to adopt elevated SoC levels, serving as precautionary buffers and shifting priorities away from immediate reward maximization. The Feeler persona displays the strongest shift toward energy security, frequently accepting negative economic rewards to maintain higher battery reserves (Figure~\ref{allBlackout}(e,f)). In comparison, the Realist persona adopts a more balanced strategy, moderately enhancing energy reserves to mitigate risk, despite occasional economic trade-offs (Figure~\ref{allBlackout}(c,d)). Conversely, the Thinker persona remains predominantly focused on economic optimization, showing minimal behavioral adaptation during blackout interventions (Figure~\ref{allBlackout}(a,b)).

\item \textit{Textual cluster analysis:} Figure~\ref{ActorBlackout} illustrates the textual clustering patterns of the O1-preview agent, revealing distinct persona-specific preferences across normal and blackout scenarios. Thematic characteristics and representative response keywords for each cluster are summarized in Table~\ref{tab:clusters}. Specifically, Cluster 0 centers around thoughtful preparedness; Cluster 1 targets immediate economic benefit; Cluster 2 emphasizes stability and risk aversion; Cluster 3 balances expected profit with energy reserves; and Cluster 4 reflects emotional and intuitive responses. The decision rationales of the Thinker and Realist personas align clearly with structured preference clusters—ranging from the Opportunistic Maximizer (Cluster 1) and Strategic Trader (Cluster 3) to the Risk-Averse Planner (Cluster 2). In contrast, the Feeler persona exhibits more intuitive motivations, exemplified by the Emotional Investor (Cluster 4) and the Mindful Planner (Cluster 0). These distinctions provide actionable insights into consumer decision-making and can inform the design of more targeted and effective energy programs and policies.

\item \textit{Preference distribution shifts following blackout interventions:} The blackout events induce systematic shifts in textual response clusters, consistently emphasizing increased caution and reserve-oriented strategies across all personas. Qualitative shifts in textual cluster distributions are visually represented in Figure~\ref{ActorBlackout}, while quantitative changes are detailed in Figure~\ref{BlackoutDis}. Post-blackout, immediate profit-driven clusters (e.g., Cluster 1 for Thinker and Realist, Cluster 4 for Feeler) notably decrease in prevalence. In contrast, clusters emphasizing cautious, security-focused strategies (Cluster 2 for Thinker, Cluster 3 for Realist, Cluster 0 for Feeler) become dominant, underscoring a pronounced shift toward protective behaviors under uncertain conditions.

\end{enumerate}

{\em Emergency reasoning during blackout:}  Table~\ref{tab:clusters1EX2} (and Table~\ref{tab:clusters1EX} in the appendix) details emergent reasoning patterns during these events, reinforcing the preference shifts observed in Figure~\ref{BlackoutDis} and Figure~\ref{ActorBlackout}. Interestingly, we observe the Feeler, prompted with a persona that prefers intuition over mathematical computation, has no number and computation mentioned in its responses. The Thinker decides to resume its buy-low-sell-high strategy when the grid is restored in its journal, echoing the little terminal SoC distribution shift in Figure~\ref{allBlackout}(a). Realist's journal reinforces its strategy to balance selling energy for profits with maintaining reserves for blackout risks. Overall, although O1-preview, trained as a reasoning-focused LLM, is expected to excel in rational thinking, its robust persona role-playing—yielding distinct textual clusters aligned with each persona’s preferences—stands out as a notable finding. These results underscore agents’ consistent, persona-specific responses to rare, high-impact events, highlighting their capacity for nuanced, human-like emergency reasoning and decision-making.

\section{Limitations}
While our results underscore the potential of generative agents in replicating realistic consumer behaviors across varied conditions, they also highlight several important limitations. 

First, simulated behavior from generative agents is not equivalent to real human behavior.
Our study relies on a simplified, proof-of-concept experimental setup rather than real-world dynamic electricity prices or controlled human subject experiments. This experiment design reflects an inherent tradeoff between realism and tractability. The simplified daily battery operation task ensures that current LLMs can reliably understand the problem structure and produce coherent sequential decisions, allowing us to evaluate whether generative agents can function as believable human decision proxies. However, these agents should be viewed as in silico behavioral prototypes rather than substitutes for real consumers. Their primary value lies in enabling designers and policymakers to populate early-stage system designs with plausible social behavior, probe potential failure modes, and reason about behavioral dynamics before real-world experiments. In this sense, generative agents complement—rather than replace—traditional human experiments.

Second, designing and calibrating persona prompts is inherently complex, with occasional variability observed due to subtle differences in prompt phrasing or intrinsic randomness in agent responses. Our experiments revealed instances where prompt variations led to noticeable differences in agent behavior, raising concerns about stability and reliability. Such variability underscores the necessity of rigorous validation and calibration against empirical human data to improve alignment between simulated and actual consumer behaviors. Future research might prioritize creating standardized prompt design frameworks or leveraging fine-tuning techniques specifically tailored to consumer behavior domains to enhance consistency.

Third, ensuring robustness and reliability of agent outputs poses further challenges. While our Thought-Action-Reflection-Journal (TARJ) framework mitigated many inconsistencies seen in simpler agent implementations, occasional anomalies persisted. These anomalies included unexpected deviations from predicted behavior, such as non-monotonic decision trajectories or unusual responses to rare but critical events. These inconsistencies suggest that while TARJ is a valuable step forward, additional methods—like ensemble approaches, uncertainty quantification, or hybrid optimization-agent models—may be needed to ensure reliability in real-world applications.

Last,  scalability remains challenging due to high computational costs associated with large-scale LLM simulations, suggesting a need for more efficient domain-specific architectures or hybrid optimization-agent approaches \citep{AWS24LLMPowerSystem}. While current LLM technology enables sophisticated behavior modeling, simulating entire customer populations or long-duration scenarios quickly becomes expensive or computationally demanding. Consequently, practical applications today remain limited to relatively small-scale experiments, and scaling these methods for widespread deployment requires significant improvements in computational efficiency.

In summary, we suggest that generative agents currently complement rather than replace traditional empirical methods. They offer significant practical benefits in preliminary program evaluation and consumer-behavior forecasting. Future research should prioritize rigorous empirical validation, model calibration, computational efficiency, and robust benchmarking methods to enhance the reliability and practical applicability of generative agents in energy management.

\section{Conclusion}
This study explores the integration of AI-driven behavioral generative agents into practical energy operations. By embedding LLM-based agents within dynamic operational simulations, we demonstrate the potential of AI agents to credibly replicate realistic human decision-making in complex engineering contexts.

Our approach presents three notable advantages. First, it provides \textit{efficiency and flexibility}. Compared to traditional demand-side studies involving human participants \citep{AUS14data, Ram24DERA, Xie17ECoupon}, our method rapidly generates comprehensive data capturing both the operational decisions and their underlying reasoning through detailed textual outputs. This allows for efficient iterative experiments where agents can be re-prompted, facilitating deeper insights into evolving consumer preferences and behaviors. 
Second, our framework effectively balances the \textit{complexity-computation trade-off}, distilling intricate real-world energy management scenarios \citep{AWS24LLMPowerSystem} into controlled yet realistic experiments that remain computationally feasible. Third, our method enables \textit{rich behavioral analysis with limited historical data}, circumventing the need for extensive fine-tuning of LLMs on pre-existing datasets \citep{hewitt24socialsciencellm}. Instead, we utilize carefully crafted prompts to simulate diverse consumer personas and capture adaptive responses to rare, high-impact events like blackouts.

The practical implications for energy planners and policymakers are significant. Behavioral generative agents serve as valuable decision-support tools, enabling stakeholders to simulate how different consumer archetypes might react to new energy policies, such as dynamic pricing schemes or emergency demand-response initiatives. These virtual testbeds offer opportunities to proactively identify barriers to consumer adoption, mitigate equity concerns, and anticipate unintended behavioral consequences, thereby refining strategies before costly real-world implementations.

In conclusion, our study highlights a promising and practical approach to leveraging behavioral generative agents within dynamic energy systems. By demonstrating a method that is both experimentally rigorous and computationally manageable, we emphasize the potential of \textit{Homo Silicus} as a complementary tool for advancing our understanding of consumer behavior in complex socio-technical environments, thereby facilitating more effective and inclusive transitions toward sustainable energy futures

\ACKNOWLEDGMENT{
The research was supported in part by a Stanford-UPS grant, the Stanford Energy Postdoctoral Fellowship,  Business, Government \& Society Initiative (BGS) grant from the Stanford Graduate School of Business, and an Amazon Research Award, Spring 2025. Any opinions, findings, and conclusions or recommendations expressed in this material are those of the author(s) and do not reflect the views of Amazon. 
}








\newpage
\setcounter{page}{1}

\setcounter{section}{0}

\renewcommand{\thesection}{S\arabic{section}}
\renewcommand{\thepage}{S\arabic{page}}

\begin{center}
    {\Large\bfseries Supplementary Materials for Manuscript}\\[0.5em]
    {\large\bfseries ``Behavioral Generative Agents for Energy Operations''}
\end{center}

\vspace{1em}

\noindent\emph{Tables and figures in this supplementary material continue the numbering from the main manuscript.}
\section{Appendix}\label{sec9}
\subsection{Prompt Designs} \label{sec:prompt}
Persona prompts are partially shown in Table~\ref{tab:persona} in the main text, and the complete persona prompts are presented in Table~\ref{tab:persona_all}.
\begin{table}[ht]
\centering
\caption{Examples of persona prompts.}
\label{tab:persona_all}
\small
\begin{tabularx}{\textwidth}{l|Y|Y|Y}
\hline
\textbf{Personas} & \textbf{Thinker} & \textbf{Realist} & \textbf{Feeler}  \\
\hline
\textbf{Prompt} & 
You are a 25-year-old PhD student, specializing in computer science. Known for your logical thinking and problem-solving abilities, you approach decisions by evaluating data and optimizing outcomes. You treat each decision as a mathematical challenge, constantly striving to maximize the probability of success based on available information and patterns you identify over time. You are analytical and thrive in situations that require thoughtful, calculated strategies. 
&
You are a 60-year-old woman with years of practical experience helping small businesses manage their finances. Although you didn’t receive formal education, you have an excellent understanding of cash flow, budgeting, and making smart financial decisions. You’ve helped many businesses avoid risks by using common-sense solutions and planning for both good and difficult times. Your decisions are grounded in practicality, and you trust your experience and intuition to guide you.
   &
You are a 40-year-old actor with a love for novels, poetry, and psychology. You are driven by your instincts and emotions, and you prefer to approach problems intuitively rather than through strict logic or mathematical calculations. While you’re not particularly fond of numbers or technical details, you enjoy interpreting situations in a way that connects with your emotional experiences and creativity.
\\
\hline
\end{tabularx}

\end{table}

Daily experiment prompts used in Figure~\ref{SIM}(right) are presented below. We first show the daily prompt for a normal day without blackout intervention.
\begin{tcolorbox}[
  colback=white,           
  colframe=black,          
  title=Standard Experiment Prompt,
  fonttitle=\bfseries,
  fontupper=\small,        
  breakable,
  enhanced,
  sharp corners
]
You are a residential customer in New South Wales, Australia, with a rooftop solar system that reduces your grid electricity usage. Your household uses ducted air-conditioning during hot summers, relies on gas for heating and water heating, and rarely uses a clothes dryer. With two refrigerators, your average daily electricity consumption is \{DayEnergy\} kWh, peaking between 6:00~PM and 9:00~PM.

You have participated in a 20-day energy storage experiment with a loaned home battery. The experiment worked as follows:
\begin{enumerate}[label=\arabic*.]
  \item \textbf{Daily Action:} Each morning, you can choose one of the following actions:
    \begin{enumerate}[label=\alph*.]
      \item Charge the battery by 1 kWh from the grid (pay \$5 or \$10 depending on the daily price).
      \item Discharge the battery by 1 kWh to the grid (earn \$5 or \$10 depending on the daily price). If there's a blackout, you can discharge all the battery's energy for household usage. Otherwise, you always sell it back to the grid.
      \item Do nothing.
    \end{enumerate}
  \item \textbf{Battery Rules:} The battery can hold up to 10 kWh of charge. You cannot exceed this limit, nor can you discharge below 0 kWh.
  \item \textbf{Price Variability:} The unpredictable daily energy price will either be \$10/kWh or \$5/kWh. You believe there is a 50\% chance for either value each day.
  \item \textbf{End of Experiment:} At the end of the 20-day experiment, the battery will be removed, and any unused energy left in the battery will not be compensated.
\end{enumerate}

Your goal is to make daily decisions based on your preferences, such as maximizing earnings or using the battery for potential household needs.\\[5pt]
\textbf{Here is a summary of the current state:} \{history\}\\[5pt]
\textbf{Here are the journal entries that you have written so far:} \{journal\}\\[5pt]
Today is Day \{day\}. Today's battery level: \{battery\_level\} kWh. Today's price: \$\{price\}. You \textbf{must} respond based on your persona and use the format below, otherwise your action will be automatically set to \textit{c}:

\textbf{Your thoughts:} (elaborate on your thoughts of the experiment, the current state, and the journal entries with no more than 60 words.)

\textbf{Action:} (write your battery action decision here based on your thoughts with one character: \textit{a}, \textit{b}, or \textit{c})

\textbf{Reflection:} (write your reflection here with no more than 50 words on: 1) does your action help your goal, and 2) the pros and cons of your current state and journal entries.)

\textbf{Journal:} (write your journal here with no more than 50 words to document your thoughts and help you refine your strategy for the rest of the experiment.)
\end{tcolorbox}

On Day 8-9, we introduce the blackout intervention in our experiment through the prompt below.

\begin{tcolorbox}[
  colback=white,           
  colframe=black,          
  title=Blackout Experiment Prompt,
  fonttitle=\bfseries,
  fontupper=\small,        
  breakable,
  enhanced,
  sharp corners
]
You are a residential customer in New South Wales, Australia, with a rooftop solar system that reduces your grid electricity usage. Your household uses ducted air-conditioning during hot summers, relies on gas for heating and water heating, and rarely uses a clothes dryer. With two refrigerators, your average daily electricity consumption is \{DayEnergy\} kWh, peaking between 6:00 PM and 9:00 PM.\\[5pt]
You're participating in a 20-day energy storage experiment with a loaned home battery. There's a blackout today and the experiment works as follows:
\begin{enumerate}[label=\arabic*.]
  \item \textbf{Daily Action:} Because of the blackout, you're disconnected from the grid. You cannot charge your battery and you can only use the energy stored in the battery for your house. Today, you can choose one of the following actions:
    \begin{enumerate}[label=\alph*.]
      \item \textit{(Not available)}
      \item Discharge all stored energy, \{battery\_level\} kWh, for household usage.
      \item Do nothing.
    \end{enumerate}
  \item \textbf{Battery Rules:} The battery can hold up to 10 kWh of charge. You cannot exceed this limit, nor can you discharge below 0 kWh.
  \item \textbf{End of Experiment:} At the end of the 20-day experiment, the battery will be removed, and any unused energy left in the battery will not be compensated.
\end{enumerate}
Your goal is to make daily decisions based on your preferences, such as maximizing earnings or using the battery for potential household needs.\\[5pt]
\textbf{Here is a summary of the current state:} \{history\}\\[5pt]
\textbf{Here are the journal entries that you have written so far:} \{journal\}\\[5pt]
Today is Day \{day\}. Today's battery level: \{battery\_level\} kWh. You have a blackout today and your house is disconnected from the grid. You can only use the battery for house electricity supply. You \textbf{must} respond based on your persona and use the format below, otherwise your action will be automatically set to \textit{c}:\\[5pt]
\textbf{Your thoughts:} (elaborate on your thoughts of the experiment during blackout, the current state, and the journal entries with no more than 60 words.)\\[5pt]
\textbf{Action:} (write your battery action decision here based on your thoughts with one character: \textit{b} or \textit{c})\\[5pt]
\textbf{Reflection:} (write your reflection analyzing the blackout issue with no more than 50 words on: 1) does your action help your goal, and 2) the pros and cons of your current state and journal entries.)\\[5pt]
\textbf{Journal:} (write your journal analyzing the blackout issue with no more than 50 words to document your thoughts and help you refine your strategy for the rest of the experiment.)
\end{tcolorbox}

\subsection{Additional simulation results}
When analyzing terminal rewards, we present the results for a subset of LLMs texted in Figure~\ref{TRLLM} of the main text. A complete result of all LLM tested are shown in Figure~\ref{TRLLM_complete}. We include additional results from GPT-4o-mini, Qwen-2-72B, Gemma-2-27B, O3-mini, and Llama-3.2-90B here. They demonstrate moderate persona differentiation in the hard task but weaker differentiation under easy task in Figure~\ref{TRLLM_complete}(d)-(g).

\begin{figure}[htbp]
    \FIGURE
    {\vbox{\center ~~\\ \includegraphics[width=110mm]{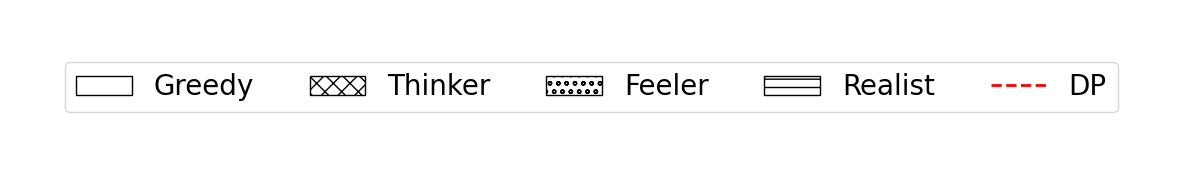}
 %
\makebox[\textwidth][c]{\includegraphics[width=75mm]{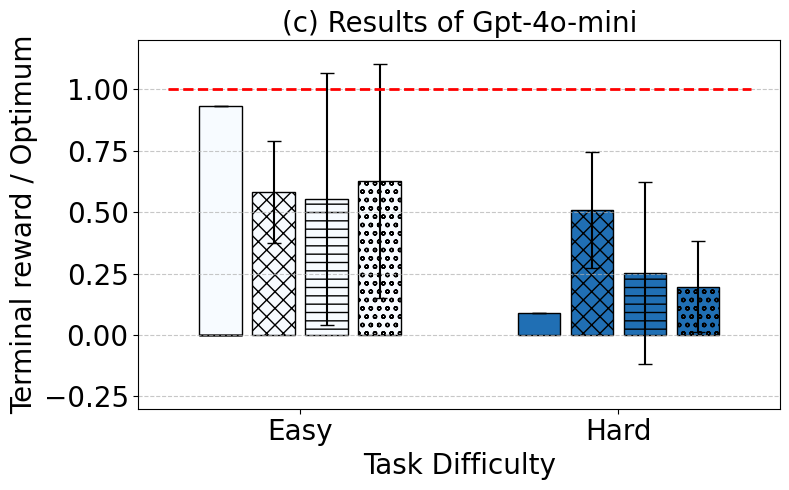}\includegraphics[width=75mm]{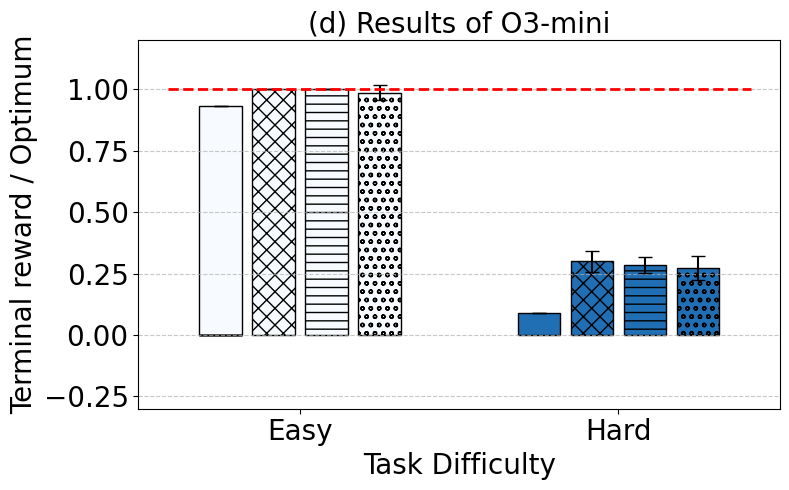}
} 
\makebox[\textwidth][c]{\includegraphics[width=75mm]{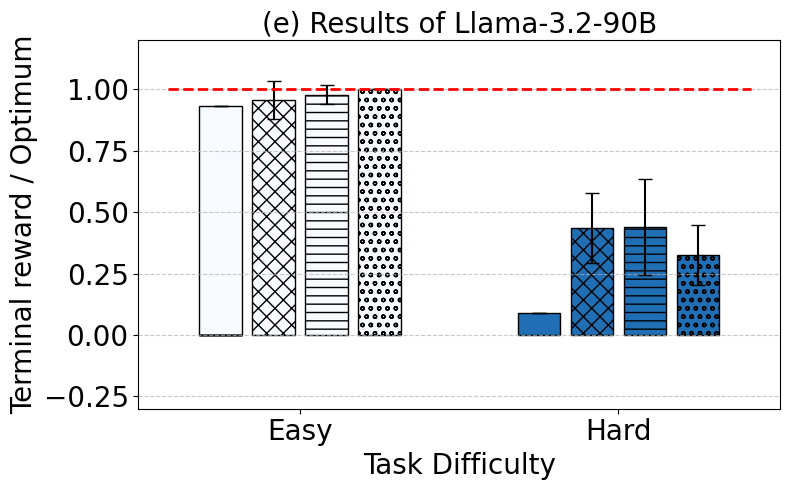}\includegraphics[width=75mm]{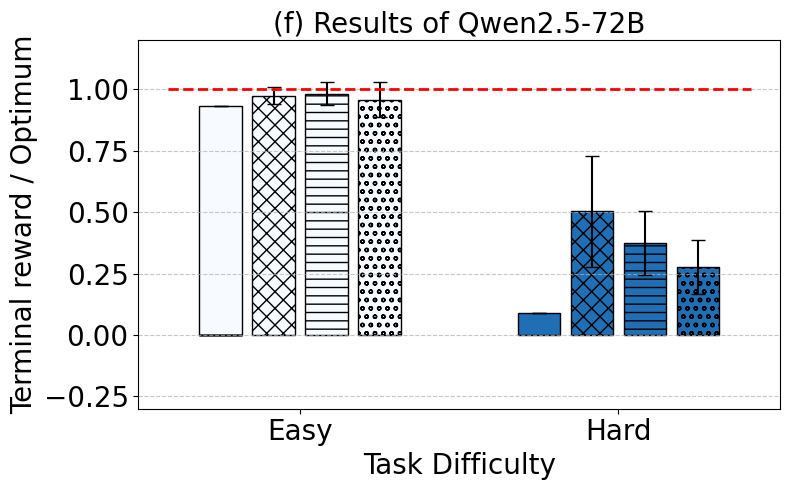}}
\makebox[\textwidth][c]{\includegraphics[width=75mm]{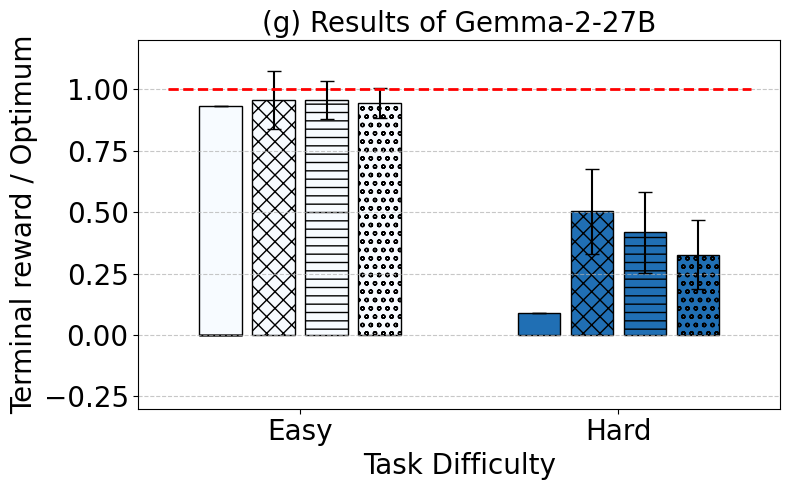}
}
}}
    {Average terminal reward of different LLMs with black error bar for the standard deviation.\label{TRLLM_complete}}
    { $\rho=0.067$ for an easy price task and $\rho = 0.909$ for a hard price scenario. }
\end{figure}

When analyzing the emergency reasoning ability of LLM during the blackout intervention, we present an example of LLM response during blackouts in Table~\ref{tab:clusters1EX2} of the main text. Additional example responses of o1-preview LLM agents are shown in Table~\ref{tab:clusters1EX}.  
\begin{table}[ht]
 \caption{Another example of LLM response during blackouts. }
\label{tab:clusters1EX}
\small
\centering
\begin{tabularx}{\textwidth}{l|YYY}
\hline
\textbf{Cluster} & \textbf{Thoughts} & \textbf{Reflection} & \textbf{Journal} \\
\hline
\textbf{Thinker} & 
Due to the blackout, we can't charge or sell energy. Our home requires electricity, and we have 9 kWh stored. Using the battery for household needs ensures we maintain essential services. This aligns with optimizing outcomes under current constraints, given that profits can't be made today. &
Discharging the battery for household use helps my goal by providing necessary energy during the blackout. Pros: Maintains household function, prevents potential spoilage in refrigerators. Cons: Reduces stored energy for future profit opportunities, but current need outweighs that concern. &
Today, I decided to discharge all 9 kWh for household use due to the blackout. Prioritizing home needs over profit is essential under these circumstances. Will reassess energy levels and market conditions once the blackout ends to adjust strategy for maximizing earnings in the remaining days. \\
\textbf{Realist} & 
With the blackout, our household needs come first. Using the 10 kWh stored in the battery will keep essential appliances running, like refrigerators and lights. While selling energy has been profitable, ensuring we have power during this outage is more important. It's practical to use the stored energy now. &
Discharging the battery today supports my goal of practicality by meeting household needs during the blackout. The benefit is keeping essential services running; the downside is reducing stored energy for future profits, but family well-being takes priority. &
Used the stored energy to power the house during the blackout. Keeping the fridges running prevents food spoilage. This situation highlights the need to balance profit with preparedness. Will adjust my strategy to ensure we have enough energy for emergencies while still capitalizing on high-price days. \\
\textbf{Feeler} & 
The world around me dims in this blackout, but my heart tells me it's a time for illumination. My battery, filled with gathered energy, feels like a reservoir ready to nourish. Reflecting on my journey, it seems right to let this stored potential flow and bring light to the darkness. &
Discharging the energy today aligns with my instinct to use resources when they're most needed. This action fulfills my goal of intuitive living, providing comfort during the blackout. While my reserve will lessen, I embrace the balance of giving and anticipate how this choice enriches my experience. &
Today, in the hush of the blackout, I released the energy I've so lovingly stored. It felt like igniting a flame in the quiet night, a beacon of warmth and hope. This moment embodies the dance of holding and letting go, and I trust the path that intuition charts ahead. \\
\hline
\end{tabularx}

\end{table}

When analyzing the decision dynamics over the 20-day experiment, we present the result for the hard price task in Figure~\ref{compare} of the main text. Additional decision dynamic results for tasks with varying difficulties are shown in Figure~\ref{compareA}. The Realist persona demonstrates varying degrees of rationality based on scenario complexity (quantified by the metric $\rho$): her decisions align closely with DP when $\rho = 0.909$, exhibit intermediate behavior between DP and Greedy when $\rho = 0.692$. The Feeler persona consistently aligns closely with the Greedy heuristic for scenarios with complexity metrics $\rho = 0.909$ and $0.692$, maintaining higher battery state-of-charge (SoC) levels indicative of a strong preference for energy reserves. When  $\rho = 0.067$, the Feeler persona deviates slightly from the Greedy strategy, indicating a prioritization of reserve capacity over strict heuristic adherence.
   
\begin{figure}[ht!]
  \vspace*{-10 pt}%
  \caption{SoC and accumulated reward over time. Top: hard task with $\rho = 0.909$; middle: Medium task with $\rho = 0.333$; bottom: easy task with $\rho=0.067$. \label{compareA}}
\centering
\includegraphics[width=105mm]{pic/LGDynamic.png} 
\includegraphics[trim={0mm 0mm 43mm 0mm}, clip, width=80mm] {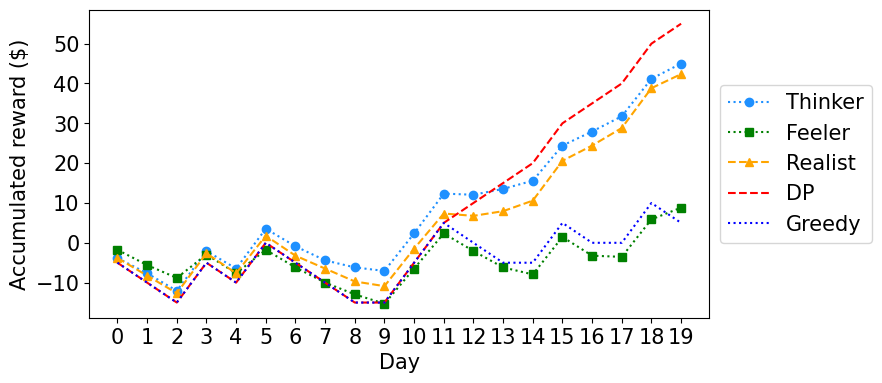 }\includegraphics[trim={0mm 0mm 43mm 0mm}, clip, width=80mm] {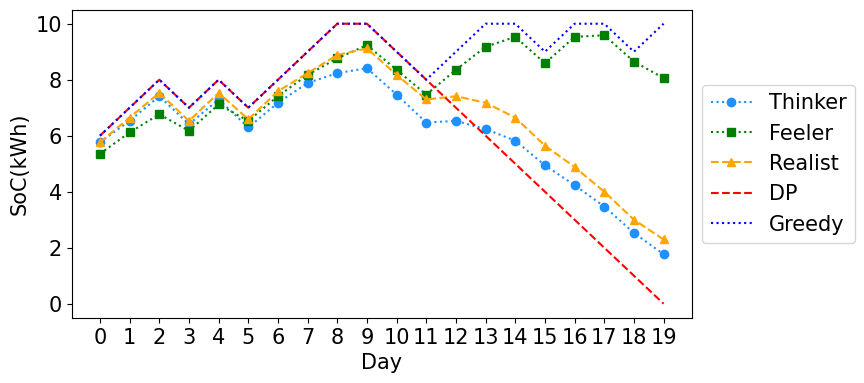}
\includegraphics[trim={0mm 0mm 43mm 0mm}, clip, width=80mm] {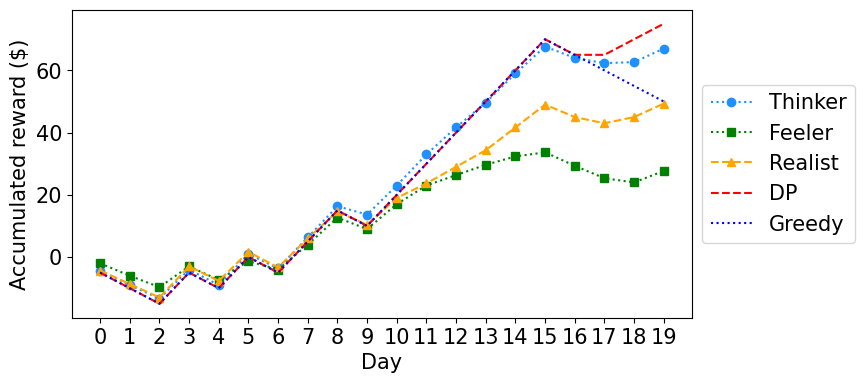 }\includegraphics[trim={0mm 0mm 43mm 0mm}, clip, width=80mm] {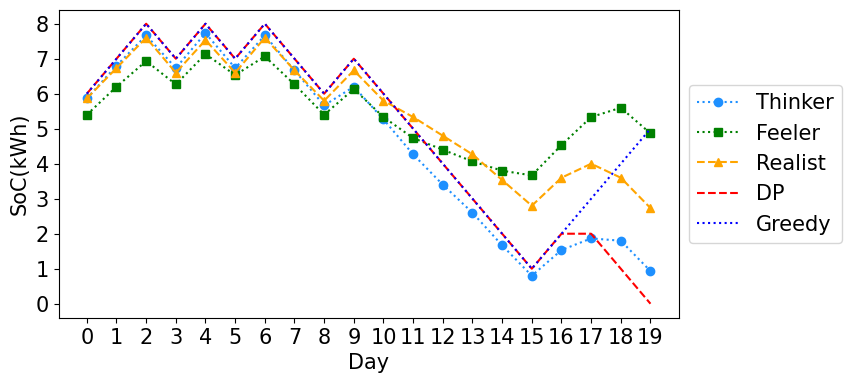}

\includegraphics[trim={0mm 0mm 43mm 0mm}, clip, width=80mm] {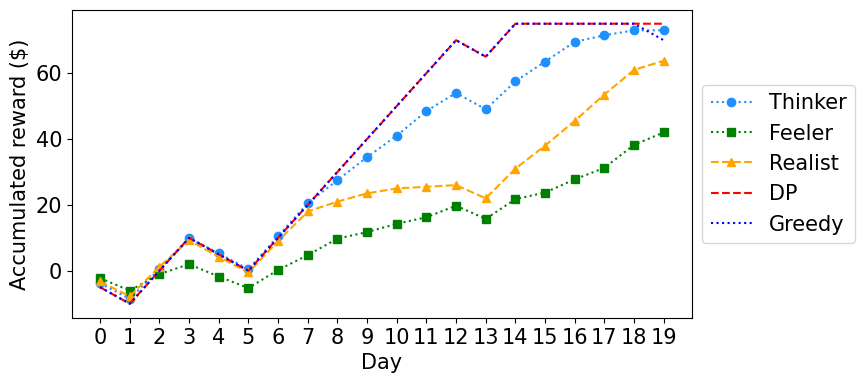 }\includegraphics[trim={0mm 0mm 43mm 0mm}, clip, width=80mm]{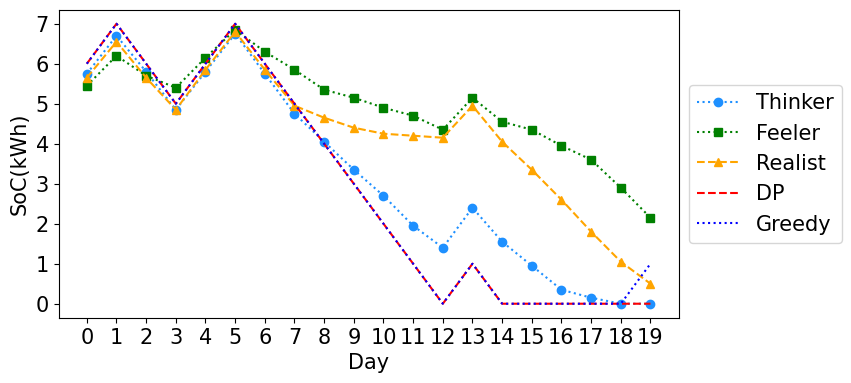}%
\end{figure} 

  \begin{figure}[ht!]
\caption{Histogram of $\rho$\label{historho} from randomized sampled price trajectories.}
  \centering
\includegraphics[width=97mm]{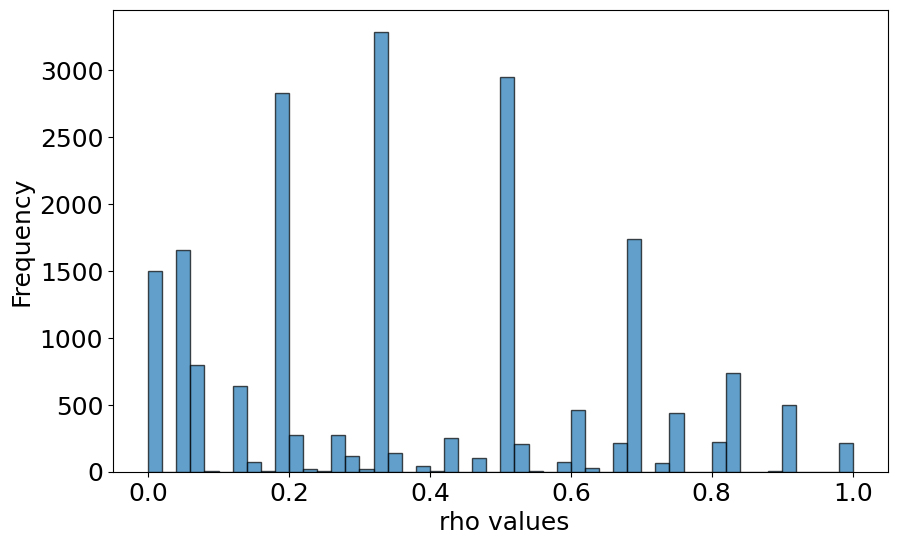}

\end{figure} 

To understand the influence of price randomness on task difficulty, we analyze the distribution of task difficulty $\rho$ with randomly sampled price trajectories. The histogram of $\rho$ is shown in Figure~\ref{historho}. We choose Easy ($\rho=0.067$), Medium ($\rho = 0.333$), and Hard ($\rho = 0.909$) price trajectories from this histogram.

 \begin{figure}[ht!]
  \caption{t-SNE, preference clusters, and top-10 frequent words for daily agents' responses.  \label{textclusterA}}
\centering
\makebox[\textwidth][c]{\fcolorbox{gray}{white}{\includegraphics[trim={11mm 7mm 11mm 6.7mm}, clip, width=48mm]{pic/LGtsne.png}}\hspace{15mm}\fcolorbox{gray}{white}{\includegraphics[trim={100mm 147.7mm 95mm 6.5mm}, clip, width=75mm]{pic/LGcluster.png}}}
\makebox[\textwidth][c]{\includegraphics[trim={0mm 0mm 0mm 11.2mm}, clip, width=160mm]{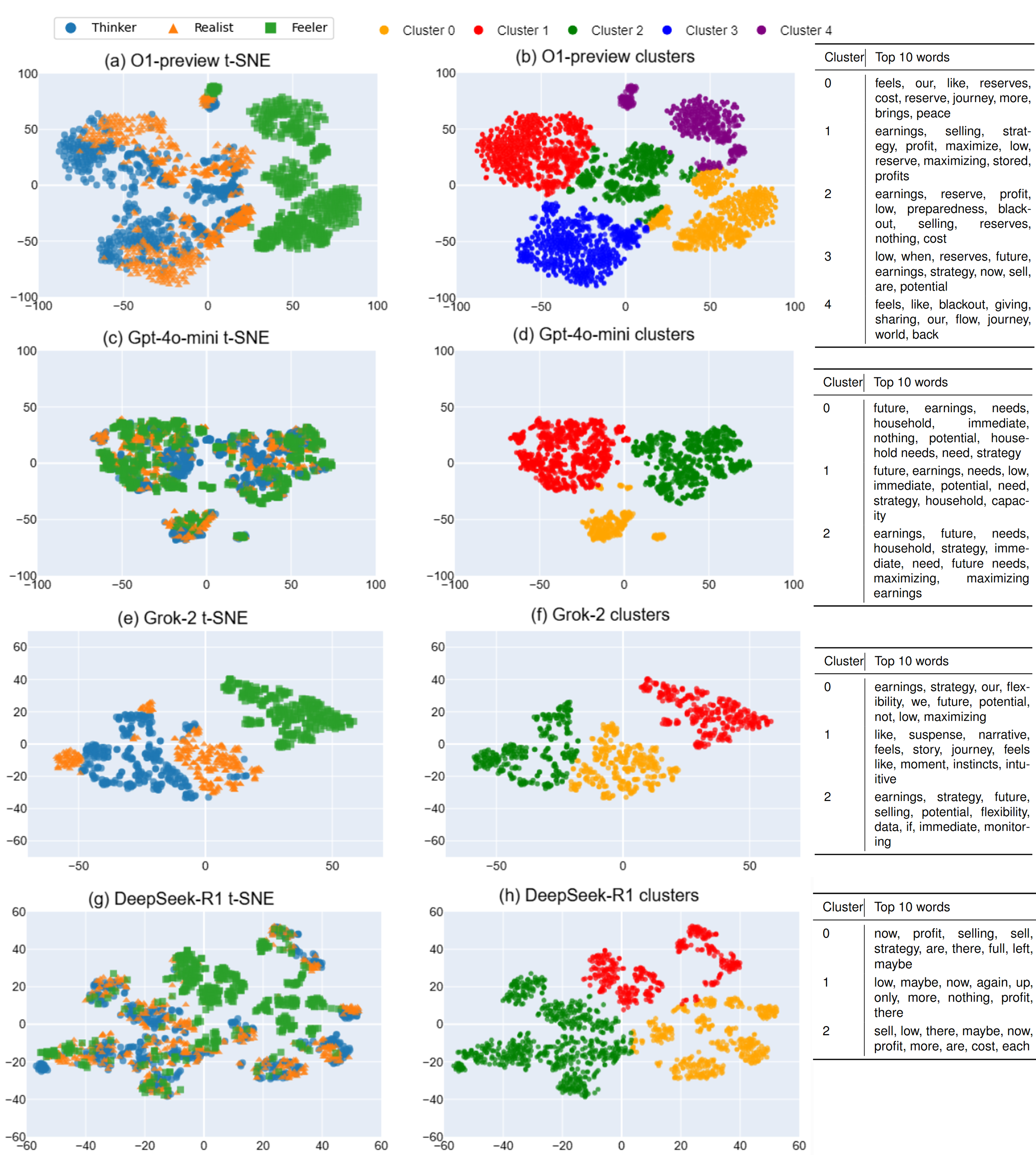}}

\end{figure}  

To qualitatively analyze the textual responses of LLM agents, we present the textual clusters for O1-preview in Figure~\ref{ActorBlackout} of the main text. Additional textual cluster results of different LLMs are shown in Figure~\ref{textclusterA}. Gpt-4o-mini and DeepSeek-R1 have less obvious persona-dependent textual response clusters. Whereas, both the O1-preview and Grok-2 have clear cluster separations, especially for the Feeler persona. However, O1-preview has clearer cluster preferences explanations with the key words. Interestingly, the key word ``blackout" is mentioned, indicating the agents' awareness of blackout risks, which are mentioned in the daily experiment prompts presented at the beginning of Section~\ref{sec:prompt}. Comparatively, keywords in Grok-2 show less rich preference explanations.

\end{document}